\documentclass[11pt,final]{article}

\usepackage[final]{acl}
\usepackage{amsmath}
\usepackage{times}
\usepackage{latexsym}
\usepackage{booktabs}
\usepackage{colortbl} 
\usepackage{multirow}
\usepackage{tabularx}
\usepackage{makecell}
\usepackage{amssymb}

\usepackage[T1]{fontenc}

\usepackage[utf8]{inputenc}

\usepackage{microtype}

\usepackage{inconsolata}

\usepackage{graphicx}

\usepackage[T1]{fontenc}
\usepackage[utf8]{inputenc}
\usepackage{amsmath}
\usepackage{microtype}
\usepackage{inconsolata}
\usepackage{graphicx}

\title{Language-Informed Flow Matching for Trend-Guided Structure-Based 3D Molecular Generation}

\author{
\textbf{Tianyu Gao}\textsuperscript{1,*} \quad
\textbf{Zhikai Su}\textsuperscript{1,*} \quad
\textbf{Jiashu Li}\textsuperscript{1} \quad
\textbf{Wenjun Gao}\textsuperscript{1}
\\
\textbf{Zichuan Ying}\textsuperscript{2} \quad
\textbf{Zhe Zhao}\textsuperscript{3} \quad
\textbf{Fei Zhang}\textsuperscript{1,\textdagger} \quad
\textbf{Ye Wei}\textsuperscript{1,\textdagger}
\\[2mm]
\textsuperscript{1}City University of Hong Kong \quad
\textsuperscript{2}The University of Hong Kong \quad
\textsuperscript{3}Stanford University
\\[1mm]
\small{
\textsuperscript{*}Equal contribution.
\quad
\textsuperscript{\textdagger}Corresponding authors.
}
\\[1mm]
\small{
\texttt{feizhang010518@gmail.com}
\quad
\texttt{ye.wei@cityu.edu.hk}
}
}

\begin{document}
\maketitle
\begin{abstract}
Structure-based drug design (SBDD) requires ligands that satisfy both 3D target affinity and 1D chemical validity. Existing controllable generation methods often rely on task-specific fine-tuning or externally imposed sampling-time guidance, adding cost and potentially conflicting with evolving 3D geometric constraints. We propose LiFT, a language-informed cross-modal framework built on Flow Matching for trend-guided 3D molecular generation across both de novo design and scaffold hopping. LiFT uses a ``Sense-Evolve-Assemble'' agent to generate target-aware SMILES as intermediate chemical conditions, from which a pre-trained chemical foundation model extracts continuous semantic priors. These priors are integrated into geometric generation through a lightweight semantic projector with zero-initialized adaptive normalization for stable cross-modal conditioning. We further introduce a Self-Conditioned Decoupled Router (SCDR), which modulates the velocity field according to intermediate structural states during ODE integration. Experiments on CrossDocked2020 show that LiFT achieves competitive distribution matching while improving medicinal chemistry metrics and maintaining competitive structural validity under task-steering settings without additional generator fine-tuning. Our results suggest that language-derived chemical priors provide effective trend-level guidance for 3D molecular generation. Code and released artifacts are available at \url{https://github.com/kasurl/LiFT}.
\end{abstract}

\section{Introduction}

Structure-based drug design (SBDD) accelerates therapeutic discovery by designing small-molecule ligands that selectively bind disease-associated protein targets \citep{Anderson2003ThePO, zhang2024geometricdeeplearningstructurebased}. Computationally, this objective requires simultaneously satisfying two constraints: \textit{3D physicochemical complementarity} within the binding pocket and \textit{1D chemical validity} \citep{Buttenschoen_2024, lin2024cbgbenchblankproteinmoleculecomplex}. Over-optimizing geometric affinity at the expense of chemical viability, or vice versa, often produces non-synthesizable or inactive molecules \citep{jiang2025chem3dllm3dmultimodallarge}. Consequently, a central challenge in SBDD is balancing 3D spatial interactions with valid chemical structures across the vast molecular space \cite{Walters2020AssessingTI}.

Purely 3D SBDD models capture target-pocket geometry, but discrete chemical constraints such as valency, ring topology, and medicinal-chemistry preferences are often only implicit. Conversely, LLM-based molecular generators can exploit chemical knowledge from SMILES and natural-language descriptions, but standalone 1D proposals cannot ensure target-specific 3D compatibility. Recent 1D--3D molecular generators such as \citet{liu2025nextmol3ddiffusionmeets} and \citet{chen2025molsculptsculpting3dmolecular} demonstrate the promise of integrating chemical priors with geometric generation, yet they mainly transfer molecular-string representations into diffusion-style generators rather than addressing how human task preferences, pocket context, and property trade-offs can be translated into semantic conditions for pocket-aware 3D generation. They are also largely pocket-free, and their cross-modal conditioning is typically fixed rather than adapted to the evolving 3D state. In SBDD, such semantic--geometric mismatch can induce steric conflicts and reduce structural validity, motivating state-aware semantic guidance during generation.

To address these limitations, we propose LiFT, a language-informed cross-modal framework built on flow matching for trend-guided 3D target-aware molecular generation. LiFT asks whether molecular strings generated from human-readable design preferences can serve as reusable semantic priors for downstream geometric generation, rather than merely acting as final molecule proposals. A ``Sense-Evolve-Assemble'' LLM agent with Pocket-of-Thought reasoning converts pocket information, task preferences, and optional references into target-aware SMILES conditions. A frozen chemical foundation encoder then maps these conditions into dense semantic latents, which are integrated into an ODE-governed flow field through zero-initialized lightweight modulation. To coordinate semantic trends with evolving spatial configurations, we further introduce the Self-Conditioned Decoupled Router (SCDR), which adapts semantic influence according to invariant summaries of intermediate structural states while preserving the $SO(3)$-equivariant geometry of the pocket-centered backbone \citep{satorras2022enequivariantgraphneural}. This design enables property-aware steering across \textit{de novo} design and reference-guided scaffold hopping without additional generator fine-tuning.

Extensive experiments on CrossDocked2020 show that LiFT provides effective trend-level guidance while maintaining competitive distribution matching and structural validity, offering a lightweight approach to language-informed SBDD generation. Our main contributions are summarized as follows:
\begin{itemize}
\item We introduce a language-informed 1D--3D conditioning pathway for target-aware SBDD, translating human-readable design preferences and pocket context into reusable SMILES-derived semantic priors. These priors are aligned with downstream geometric generation through state-aware cross-modal modulation, enabling property-aware trend guidance without treating LLM outputs as final molecules.

\item We introduce the Self-Conditioned Decoupled Router (SCDR), coupled with zero-initialized lightweight modulation, to adapt semantic influence according to intermediate spatial states and coordinate language-derived chemical priors with geometric constraints along the integration trajectory.

\item We design an evaluation protocol that jointly measures distribution matching, property-oriented generation, and industrial chemical filter compliance. Under this protocol, LiFT shows competitive distribution matching and effective trend steering across both \textit{de novo} design and scaffold hopping without additional generator fine-tuning.

\end{itemize}

\section{Related Work}

\subsection{Generative Models in Structure-Based Drug Design}

Structure-based drug design (SBDD) aims to generate ligands that are compatible with 3D protein pockets in both geometry and chemistry. Recent 3D generative models directly operate on atomic coordinates, including spatial autoregressive models \citep{luo20223dgenerativemodelstructurebased,pmlr-v162-peng22b}, diffusion-based methods \citep{guan20233dequivariantdiffusiontargetaware,huang2024bindingadaptivediffusionmodelsstructurebased,guan2024decompdiffdiffusionmodelsdecomposed,schneuing2024structurebaseddrugdesignequivariant}, and continuous-time Flow Matching frameworks \citep{zhang2024rectifiedflowstructurebased,dunn2025flowmol3flowmatching3d,zhou2025priorguided,schneuing2025multidomain}. These approaches are well suited for modeling pocket-conditioned spatial configurations, but discrete chemical constraints such as valency, ring patterns, and medicinal-chemistry preferences are not always explicit in continuous coordinate generation.

A complementary line of work treats molecules as 1D sequences and applies language-modeling architectures to SMILES or related representations \citep{Bagal2021MolGPTMG}. Such models naturally capture chemical syntax and topological regularities, but their connection to target-specific 3D pocket geometry is less direct. Target-aware sequence models have been explored for drug design \citep{Wu2024TamGenDD}, while recent methods such as ELILLM further investigate the use of language-model representations for structure-based molecular design \citep{hu2026empoweringllmsstructurebaseddrug}. Our work follows this broader direction but treats language-derived molecular representations as intermediate semantic conditions for native pocket-conditioned 3D generation, with state-aware modulation along the flow trajectory.

\subsection{Cross-Modal Conditioning and State-Aware Modulation}

Cross-modal conditioning is commonly used to incorporate external information into generative models. Existing approaches include cross-attention or graph-language interaction modules \citep{liu2024molcamoleculargraphlanguagemodeling}, as well as FiLM- or AdaLN-style modulation in diffusion and transformer architectures \citep{perez2017filmvisualreasoninggeneral,peebles2023scalablediffusionmodelstransformers}. Recent 1D--3D molecular generation methods further use learned projection modules to connect molecular representations with 3D generation processes \citep{liu2025nextmol3ddiffusionmeets,chen2025molsculptsculpting3dmolecular}. Our work follows this direction, but uses zero-initialized lightweight modulation \citep{zhu2026unveilingsecretadalnzerodiffusion} to integrate SMILES-derived priors into the flow model with limited architectural overhead.

Frozen chemical foundation models provide a useful interface for this conditioning by encoding SMILES into continuous representations that capture chemical syntax and property-related regularities from large molecular corpora~\citep{Ross2021LargescaleCL,Soares2025AMF,Soares2025AnOF}. We use these representations as soft chemical priors for 3D generation, rather than exact structural targets. Unlike inference-time guidance methods that modify sampling or optimization trajectories with external objectives~\citep{choi2025controllable3dmoleculargeneration,10.1021/acs.jcim.6c00964,guo2025trainingfree}, our conditioning signal is incorporated into the learned velocity field.

Because a fixed semantic signal may not be equally useful throughout ODE generation, our framework uses state-aware modulation to adapt the influence of SMILES-derived priors according to intermediate ligand states while preserving the geometric backbone of the 3D flow model.

\section{Methodology}

\begin{figure*}[t] 
  \centering
  \includegraphics[width=\textwidth]{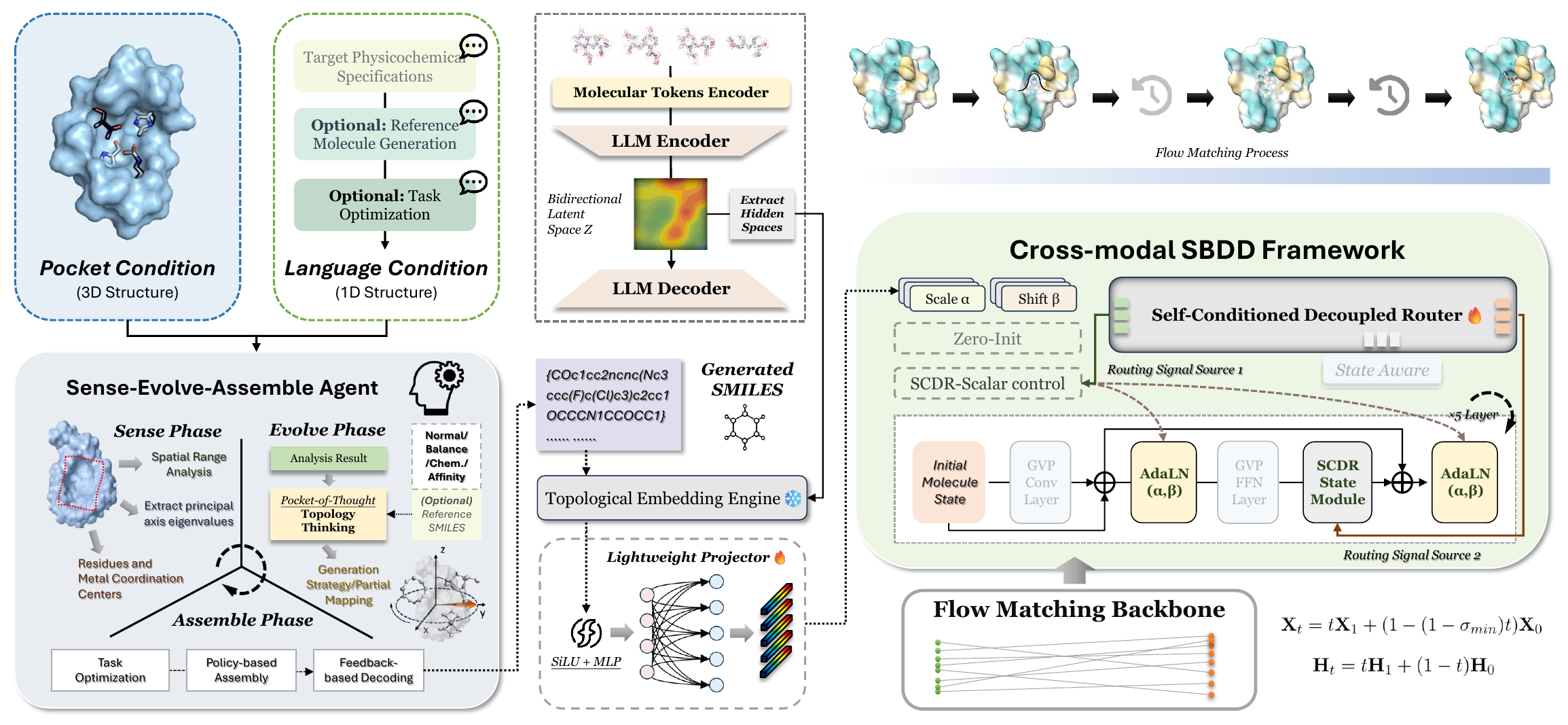} 

  \caption{\textbf{Overview of the LiFT Framework.} The left panel details the language-conditioned semantic pathway, utilizing the PoT agent, SMILES sanitization, and a foundation encoder to project sequences into dense priors. The right panel illustrates the cross-modal generation framework, injecting these semantic priors into an ODE-governed vector field with SCDR providing state-aware dynamic velocity modulation during flow matching.}
  \label{fig:framework}
\end{figure*}

We propose LiFT, a language-conditioned framework for pocket-conditioned 3D molecular generation (Figure~\ref{fig:framework}).
LiFT builds on continuous-time Flow Matching~\citep{schneuing2025multidomain}, learning a vector field that transports prior noise toward the empirical ligand distribution conditioned on a 3D protein pocket.
It consists of four stages: (i) target-aware SMILES generation with the \textit{Sense-Evolve-Assemble LLM Agent} (\S\ref{sec:agent}); (ii) semantic latent extraction from generated SMILES (\S\ref{sec:latent}); (iii) scalar-domain conditioning through a lightweight semantic projector (\S\ref{sec:projector}); and (iv) state-aware velocity-field modulation with the \textit{Self-Conditioned Decoupled Router} (\S\ref{sec:scdr}).
Detailed flow-matching derivations are provided in Appendix~\ref{appendix:math}.

\subsection{Sense-Evolve-Assemble LLM Agent}
\label{sec:agent}

To make language conditioning explicit and chemically grounded, we introduce a \textit{Sense-Evolve-Assemble} agent based on a Pocket-of-Thought (PoT) prompting strategy. Instead of directly decoding a molecule from the protein pocket, the agent first constructs a structured pocket profile, then derives task-specific SMILES candidates, and finally sanitizes the generated sequence before passing it to the 3D generator.

\begin{figure*}[t]
    \centering
    \includegraphics[width=0.85\textwidth]{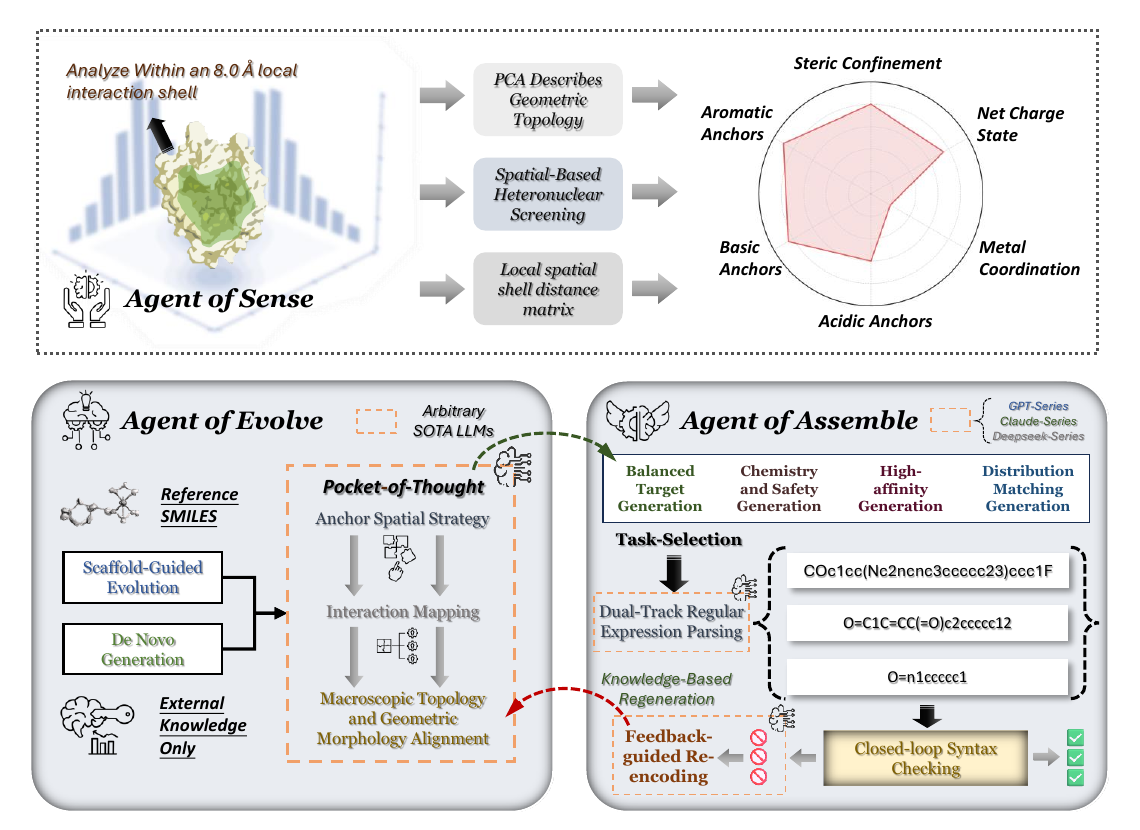} 
    \caption{Workflow of the ``Sense-Evolve-Assemble'' LLM agent. \textbf{(Top)} The \textit{Sense phase} extracts spatial geometry and physicochemical anchors within an 8.0 \AA{} pocket interaction shell to build the target profile. \textbf{(Bottom Left)} The \textit{Evolve phase} executes Pocket-of-Thought (PoT) reasoning, leveraging text specifications or reference structures for both \textit{de novo} design and scaffold hopping. \textbf{(Bottom Right)} The \textit{Assemble phase} performs rule-based sequence serialization, employing deterministic cheminformatics verification to enforce 1D topological and chemical validity.}
    \label{fig:agent}
\end{figure*}

In the \textbf{Sense} phase, the agent summarizes the local microenvironment of the pocket $\mathcal{P}$. It defines the generative region with a spatial bounding box $\mathcal{B} \in \mathbb{R}^{3 \times 2}$ and applies Principal Component Analysis (PCA) to pocket coordinates to estimate the major geometric axes. We further restrict the interaction analysis to an $8.0\,\text{\AA}$ shell around the pocket centroid, within which the agent extracts key physicochemical cues, including hydrophobic residues, acidic and basic residues, and metal coordination centers such as ZN and MG.

During the \textbf{Evolve} phase, the agent performs task-specific ligand proposal under either reference-free (\textit{de novo}) or reference-guided (\textit{scaffold hopping}) settings. Guided by PoT, it first produces a structured reasoning path rather than a raw molecule, relating candidate substructures such as rings or heteroatoms to nearby pocket features and assessing whether the proposed molecular topology is compatible with the pocket geometry. The prompt also incorporates medicinal chemistry preferences, allowing different generation modes to emphasize distribution matching, binding-oriented design, drug-likeness, or balanced property profiles.

In the \textbf{Assemble} phase, the proposed components are serialized into a raw SMILES sequence $S_{\mathrm{raw}}$. To reduce hallucinated or invalid outputs, we apply deterministic knowledge-guided decoding with cheminformatics checks for syntax errors, unmatched ring closures, and valency violations. The repaired and sanitized sequence $S_{\mathrm{smiles}}$ is then used as the 1D semantic condition for the downstream continuous 3D flow matching process.

\subsection{Semantic Latent Extraction for Cross-Modal Priming}
\label{sec:latent}

To connect discrete SMILES conditions with continuous geometric generation, we encode the agent-generated $S_{smiles}$ using SMI-TED, a chemical foundation model pre-trained on 91M molecules. The encoder maps the 1D molecular syntax and associated chemical context into a global semantic vector:
\begin{equation}
\mathbf{z}_{sem} = \text{Encoder}_{\text{SMI-TED}}(S_{smiles})
\end{equation}

We use $\mathbf{z}_{sem}$ as the cross-modal prior for the downstream flow model.
Compared with discrete token injection, this latent representation provides a compact continuous interface for conditioning the 3D velocity field with SMILES-derived chemical information.
Because SMI-TED is pre-trained on large molecular corpora, $\mathbf{z}_{sem}$ carries pharmacological priors that can bias geometric generation toward property-oriented trends without fine-tuning the generator.

\subsection{Scalar Priming via Lightweight Semantic Projector}
\label{sec:projector}

SBDD requires combining 1D symbolic descriptors ($\mathbf{z}_{sem}$) with 3D geometric configurations. We represent each ligand state as a decoupled pair of invariant scalar features and proper-rotation-equivariant vector features:
\begin{equation}
\mathbf{x} = (\mathbf{s}, \mathbf{V}) \in \mathbb{R}^{N_L \times d_s} \times \mathbb{R}^{N_L \times d_v \times 3}
\end{equation}
Within this centered frame, $SO(3)$ rotations are the geometric transformations preserved by the reported reflection-sensitive backbone. We therefore apply cross-modal priming only to the scalar domain $\mathbf{s}$. This design keeps the equivariant vectors $\mathbf{V}$ unchanged by the non-equivariant 1D signal, allowing semantic conditioning to enter the model without disrupting the geometric update rules. 

A lightweight projector $\mathcal{P}_\phi$ maps the semantic latent into affine modulation parameters:
\begin{equation}
\boldsymbol{\gamma}, \boldsymbol{\beta} = \mathcal{P}_\phi(\mathbf{z}_{sem}) = \mathbf{W}_p \cdot \sigma(\mathbf{z}_{sem}) + \mathbf{b}_p
\end{equation}
where $\boldsymbol{\gamma}, \boldsymbol{\beta} \in \mathbb{R}^{d_s}$. We employ a \textbf{zero-initialization} strategy ($\mathbf{W}_p, \mathbf{b}_p = \mathbf{0}$), so the projector initially recovers the original scalar normalization path and introduces semantic conditioning only through learned deviations. Let $\mathrm{AdaLN}_{\gamma,\beta}(\mathbf{s})=\text{LN}(\mathbf{s})\odot(1+\tanh(\boldsymbol{\gamma}))+\boldsymbol{\beta}$. The primed node state is then defined as:
\begin{equation}
\mathbf{s}_{base}=\mathrm{AdaLN}_{\gamma,\beta}(\mathbf{s}), \, \mathbf{V}_{base}=\mathbf{V}.
\end{equation}
Thus, $\mathbf{s}_{base}$ carries SMILES-derived priors before state-aware refinement by SCDR.

\subsection{Self-Conditioned Decoupled Router (SCDR)}
\label{sec:scdr}

\begin{figure*}[t]
    \centering
    \includegraphics[width=0.95\textwidth]{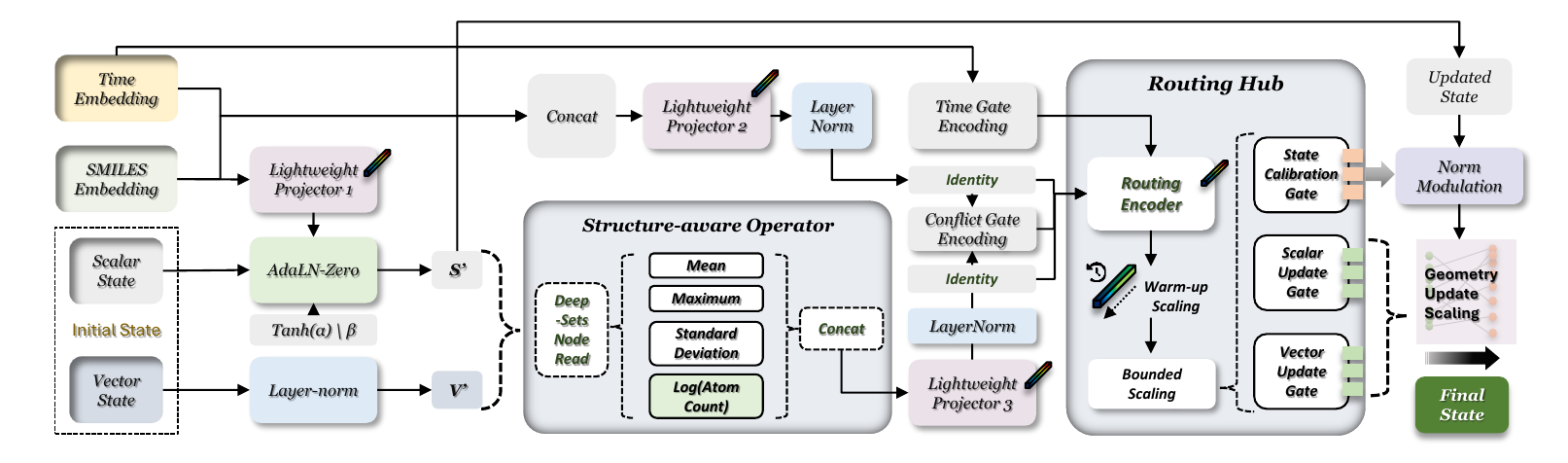}
    \caption{\textbf{Detailed architecture of the Self-Conditioned Decoupled Router (SCDR).} Temporal/SMILES priors and intermediate geometric states ($S', V'$) are contextualized and aggregated by the \textbf{Structure-aware Operator} through a multi-scale DeepSets node readout. The \textbf{Routing Module} then decodes the fused representation into three decoupled pathways: the \textit{State Calibration Gate} ($\alpha_{\mathrm{state}}^s$), the \textit{Scalar Update Gate} ($\mathbf{g}_{\mathrm{up}}^s$), and the \textit{Vector Update Gate} ($\mathbf{g}_{\mathrm{up}}^v$) for downstream geometric update scaling.}
\label{fig:scdr_architecture}
\end{figure*}

To coordinate semantic conditioning with geometric generation, we propose the \textit{Self-Conditioned Decoupled Router} (SCDR), a cross-modal routing module that adjusts semantic influence according to intermediate 3D structures, thereby reducing conflicts between 1D chemical priors and pocket geometry.

During ODE integration, SCDR summarizes the evolving ligand state through invariant structural statistics. A DeepSets encoder projects node features into $SO(3)$-invariant latents $\mathbf{z}_{node} = \phi([\mathbf{s} \parallel \|\mathbf{V}\|])$, which are aggregated by a statistical readout operator $\Psi_{\mathrm{stat}}$:
\begin{equation}
\begin{aligned}
\mathbf{z}_{\mathrm{stat}} = \big[ \, &\mathbb{E}(\mathbf{z}_{node}) \parallel \max(\mathbf{z}_{node}) \parallel \\
&\sigma(\mathbf{z}_{node}) \parallel \log(N) \, \big]
\end{aligned}
\end{equation}
where $N$ is the atom count. Combining mean, maximum, dispersion, and scale statistics, $\mathbf{z}_{\mathrm{stat}}$ provides a compact permutation-invariant summary of the current ligand configuration.

To integrate structural context with semantic conditioning, both sources are projected into a shared latent space, yielding $\mathbf{\hat{h}}_{struct}$ and $\mathbf{\hat{h}}_{cond}$, where $\mathbf{\hat{h}}_{cond}$ encodes the semantic prior $\mathbf{z}_{sem}$ and time embedding $\mathbf{t}_{emb}$. Instead of directly adding the two representations, SCDR uses a dual-gated fusion mechanism:
\begin{align}
g_{t} &= \text{Sigmoid}(\text{MLP}_{t}(\mathbf{t}_{emb})) \\
g_{c} &= \text{Sigmoid}(\text{MLP}_{c}([\mathbf{\hat{h}}_{cond} \parallel \mathbf{\hat{h}}_{struct}]))
\end{align}
The temporal gate $g_{t}$ controls how much structural feedback is used at different ODE stages, while the context gate $g_{c}$ modulates the interaction between semantic and structural representations. The unified router latent $\mathbf{h} \in \mathbb{R}^{d_h}$ is obtained by gated residual fusion:
\begin{equation}
\mathbf{h} = \text{LayerNorm}\big(\text{MLP}_{fuse}(\mathbf{\hat{h}}_{cond} + g_{t} \cdot g_{c} \cdot \mathbf{\hat{h}}_{struct})\big)
\end{equation}

\begin{table*}[t]
    \centering
    \caption{Comprehensive distribution matching benchmark on CrossDocked2020. All metrics represent the Wasserstein Distance (WD ↓) between generated molecular-property distributions and the empirical ligand distribution estimated from the 100K training complexes. The evaluation is rigorously divided into Binding Efficiency (Vina, Gnina), Medicinal Chemistry limits (QED, SA, LogP, Rotatable Bonds), and Topological Ring Systems (minimum ring frequencies $>0$, $>10$, $>100$). The best results are highlighted in \textbf{bold}, the second best are \underline{underlined}, and the third best are marked with a dagger ($^{\dagger}$).}
    \label{tab:main_results}
    \resizebox{\textwidth}{!}{
        \begin{tabular}{l cc cccc ccc}
            \toprule
            \multirow{2}{*}{\textbf{Method}} & \multicolumn{2}{c}{\textbf{Binding Eff. (WD $\downarrow$)}} & \multicolumn{4}{c}{\textbf{MedChem Properties (WD $\downarrow$)}} & \multicolumn{3}{c}{\textbf{Topological Rings (WD $\downarrow$)}} \\
            \cmidrule(lr){2-3} \cmidrule(lr){4-7} \cmidrule(lr){8-10}
            & Vina & Gnina & QED & SA & LogP & Rot. Bonds & $>0$ & $>10$ & $>100$ \\
            \midrule
            \multicolumn{10}{l}{\textit{Autoregressive Models}} \\
            \midrule
            AR \cite{luo20223dgenerativemodelstructurebased} & 0.032$^{\dagger}$ & \underline{0.020} & 0.036 & 1.066 & 1.389 & 1.370 & 0.429 & 0.420 & 0.402 \\
            Pocket2Mol \cite{pmlr-v162-peng22b} & 0.047 & 0.023 & 0.057 & \textbf{0.127} & 0.737 & 3.087 & 0.320 & 0.316 & 0.300 \\
            \midrule
            \multicolumn{10}{l}{\textit{3D Diffusion Architectures}} \\
            \midrule
            TargetDiff \cite{guan20233dequivariantdiffusiontargetaware} & 0.035 & 0.031 & 0.054 & 1.513 & 0.497$^{\dagger}$ & 0.355$^{\dagger}$ & 0.488 & 0.504 & 0.498 \\
            DecompDiff \cite{guan2024decompdiffdiffusionmodelsdecomposed} & 0.112 & 0.070 & 0.080 & 1.307 & 0.643 & 2.421 & 0.343 & 0.355 & 0.336 \\
            BindDM \cite{huang2024bindingadaptivediffusionmodelsstructurebased} & \textbf{0.018} & 0.026 & 0.025$^{\dagger}$ & 1.559 & 0.648 & 0.723 & 0.522 & 0.532 & 0.533 \\
            \midrule
            \multicolumn{10}{l}{\textit{Flow-Matching Frameworks}} \\
            \midrule
            PAFlow \cite{zhou2025priorguided} & 0.053 & 0.022$^{\dagger}$ & 0.050 & 1.664 & 2.462 & 1.392 & 0.608 & 0.623 & 0.602 \\
            DrugFlow \cite{schneuing2025multidomain}     & 0.042 & 0.024 & \underline{0.023} & 0.243$^{\dagger}$ & \underline{0.475} & \underline{0.233} & 0.139 & 0.127 & 0.098 \\
            \midrule
            \multicolumn{10}{l}{\textit{LLM-based SBDD Pipelines}} \\
            \midrule
            TamGen \cite{Wu2024TamGenDD} & 0.376 & 0.186 & 0.076 & 4.303 & 0.945 & 5.149 & 0.028$^{\dagger}$ & \underline{0.008} & 0.047$^{\dagger}$ \\
            ELILLM-Diff \cite{hu2026empoweringllmsstructurebaseddrug} & 0.377 & 0.190 & 0.048 & 5.341 & 3.424 & 1.307 & 0.521 & 0.540 & 0.527 \\
            ELILLM-Rand \cite{hu2026empoweringllmsstructurebaseddrug} & 0.376 & 0.184 & 0.074 & 4.888 & 2.152 & 2.153 & 0.382 & 0.414 & 0.406 \\
            \midrule
            \multicolumn{10}{l}{\textbf{\textit{LiFT Variants}}} \\
            \midrule
            LiFT (Ligand-Ref Embedding)     & 0.034 & \underline{0.020} & 0.054 & 0.437 & 0.954 & 0.409 & 0.260 & 0.244 & 0.207 \\
            LiFT (Norm No-Ref)        & 0.056 & 0.040 & 0.203 & 0.538 & 1.221 & 1.012 & \underline{0.027} & 0.020$^{\dagger}$ & \textbf{0.010} \\
            LiFT (Norm Ref)  & 0.041 & 0.029 & 0.083 & \underline{0.232} & \textbf{0.374} & 0.624 & 0.162 & 0.163 & 0.145 \\
            \textbf{LiFT (Balanced Ref)} & 0.044 & 0.027 & \textbf{0.020} & 0.416 & 0.762 & \textbf{0.229} & 0.192 & 0.185 & 0.165 \\
            \textbf{LiFT (Vina No-Ref)}  & \underline{0.031} & \textbf{0.019} & 0.197 & 0.568 & 1.104 & 1.400 & \textbf{0.008} & \textbf{0.004} & \underline{0.028} \\
            \bottomrule
        \end{tabular}
    }
\end{table*}

Conditioned on $\mathbf{h}$, the router branches into two pathways. In the \textbf{State Path}, the graph-level latent $\mathbf{h}_{b(i)}$ is broadcast to atom $i$ to infer a bounded scalar gate $\boldsymbol{\alpha}_{state, i}^s \in \mathbb{R}^{d_s}$:
\begin{align}
\boldsymbol{\alpha}_{state, i}^s &= \mathbf{1} + w(p) b_{a_s}(\mathbf{W}_{state}\mathbf{h}_{b(i)}) \\
\mathbf{s}_{out, i} &= \mathbf{s}_{base, i} + \text{LN}(\mathbf{s}_i) \odot (\boldsymbol{\alpha}_{state, i}^s - \mathbf{1})
\end{align}
where $\mathbf{W}_{state} \in \mathbb{R}^{d_s \times d_h}$ is a learnable projection and $b_a(\mathbf{z}) = a\mathbf{z} / (\mathbf{1} + |\mathbf{z}|)$. Setting the state scale $a_s=0.5$ bounds the gate within $[0.5, 1.5]$ for stable calibration. The vector state bypasses this modulation ($\mathbf{V}_{out, i} = \mathbf{V}_i$), preserving the pocket-centered $SO(3)$ equivariance of the geometric backbone.

Simultaneously, the \textbf{Update Path} modulates the geometric feed-forward network $\mathcal{F} = (\mathcal{F}_s, \mathcal{F}_v)$ to regulate residual updates during generation. For each channel $k \in \{s, v\}$, it infers decoupled scaling factors $\mathbf{g}_{up, i}^k$:
\begin{align}
\mathbf{g}_{up, i}^k &= \mathbf{1} + w(p) b_{a_u}(\mathbf{W}_{up}^k \mathbf{h}_{b(i)}) \\
\mathbf{x}_{next, i} &= \mathbf{x}_{out, i} + \mathbf{g}_{up, i} \odot \mathcal{F}(\mathbf{x}_{out, i})
\end{align}
where $\mathbf{W}_{up}^k$ are the corresponding learnable weights, and $\mathbf{g}_{up, i} = (\mathbf{g}_{up, i}^s, \mathbf{g}_{up, i}^v)$. A broader update scale $a_u=0.9$ permits residual scaling within $[0.1, 1.9]$. By separating scalar state calibration from residual update modulation, SCDR enables semantic priors to interact with intermediate 3D structural states while maintaining equivariant geometric updates.

\section{Experiments}

In this section, we evaluate LiFT from two complementary perspectives: distribution matching fidelity and property-oriented generation.
We report distribution matching in Sec.~\ref{sec:Comprehensive Distribution}, property-oriented results in Sec.~\ref{multiobjective}, and ablation studies in Sec.~\ref{sec:ablation_mechanics}.
Extended analyses of language-derived conditions are provided in Appendix~\ref{app:condition_analysis} and~\ref{app:alignment_suite}, while implementation details, extended metrics, and qualitative cases are provided in Appendix~\ref{appendix:reproducibility}, Appendix~\ref{appendix:viz}, and Appendix~\ref{sec:appendix_case_study}, respectively.

\subsection{Experimental Setup}
\label{sec:Experimental Setup}

\textbf{Datasets and Baselines.} We train on the refined CrossDocked2020 dataset \citep{Francoeur2020ThreeDimensionalCN} ($100$K complexes) and compare with autoregressive models \citep{pmlr-v162-peng22b,luo20223dgenerativemodelstructurebased}, 3D diffusion methods \citep{guan20233dequivariantdiffusiontargetaware,guan2024decompdiffdiffusionmodelsdecomposed,huang2024bindingadaptivediffusionmodelsstructurebased}, flow-matching frameworks \citep{schneuing2025multidomain,zhou2025priorguided}, and LLM-based SBDD pipelines \citep{Wu2024TamGenDD,hu2026empoweringllmsstructurebaseddrug}.

\textbf{Implementation Details.}
Our framework uses a 5-layer heterogeneous GVP-GNN~\citep{jing2021learning} with a frozen SMI-TED encoder~\citep{Soares2025AnOF}, trained by continuous flow matching with $T=500$ ODE steps.
Hardware, inference speed, and hyperparameters are provided in Appendix~\ref{appendix:details_hyperparams}.

\textbf{Evaluation Metrics.}
We evaluate 1D pharmacological/topological properties (QED, SA, LogP, and ChEMBL ring frequencies), 3D binding and validity (Vina/Gnina and PoseBusters), and filter compliance (REOS and RDKit alerts).
For distribution matching, Wasserstein Distance (WD) is computed against the empirical ligand distribution from the 100K training complexes.
Metric details are provided in Appendix~\ref{appendix:details_metrics}.

\textbf{Information Access.}
Reference-free variants use only pocket-derived textual specifications and task preferences, without access to ground-truth ligand SMILES, graphs, coordinates, docking poses, or ligand-derived labels.
Reference-guided variants additionally use an available ligand SMILES only as a scaffold-hopping anchor, not as an evaluation target.
Property-oriented variants change only inference-time task instructions after training the base 3D generator, without additional generator fine-tuning or external guidance models.

\begin{table*}[t]
\centering
\renewcommand{\arraystretch}{1}
\caption{Property-oriented generation without additional generator fine-tuning on CrossDocked2020. Binding efficiency metrics (Gnina $\uparrow$, Vina $\downarrow$) are reported as absolute means. MedChem metrics (QED $\uparrow$, SA $\downarrow$) reflect the mean molecular properties, and topological/safety filters (Rings $\uparrow$, PoseB./RDKit/REOS $\uparrow$) are reported as pass rates (\%). Best results are highlighted in \textbf{bold}, second best \underline{underlined}, third best $^{\dagger}$.}
\label{tab:optimization_results}
\resizebox{\textwidth}{!}{
\begin{tabular}{l l cc cc ccc ccc}
\toprule
\multirow{2}{*}{\textbf{Category}} & \multirow{2}{*}{\textbf{Method}} & \multicolumn{2}{c}{\textbf{Binding Aff.}} & \multicolumn{2}{c}{\textbf{MedChem Means}} & \multicolumn{3}{c}{\textbf{Rings Freq (\%) $\uparrow$}} & \multicolumn{3}{c}{\textbf{Filters (\%) $\uparrow$}} \\
\cmidrule(lr){3-4} \cmidrule(lr){5-6} \cmidrule(lr){7-9} \cmidrule(lr){10-12}
 & & Gnina $\uparrow$ & Vina $\downarrow$ & QED $\uparrow$ & SA $\downarrow$ & >0 & >10 & >100 & RDKit & REOS & PoseB. \\
\midrule
\multirow{7}{*}{\shortstack{3D-based\\Models}} 
 & AR \cite{luo20223dgenerativemodelstructurebased} & \underline{0.295} & -0.414$^\dagger$ & 0.509 & 4.289 & 40.99 & 36.48 & 31.09 & 52.69 & 43.17 & 59.86 \\
 & PAFlow \cite{zhou2025priorguided} & 0.262 & \textbf{-0.429} & 0.491 & 4.889 & 23.02 & 16.20 & 10.99 & 75.55 & 58.28 & 14.68 \\
 & BindDM \cite{huang2024bindingadaptivediffusionmodelsstructurebased} & 0.249 & -0.362 & 0.508 & 4.784 & 31.68 & 25.31 & 17.98 & 69.81 & 56.83 & 31.93 \\
 & DecompDiff \cite{guan2024decompdiffdiffusionmodelsdecomposed} & 0.205 & -0.267 & 0.454 & 4.532 & 49.58 & 43.01 & 37.60 & 60.27 & 44.62 & 54.92 \\
 & TargetDiff \cite{guan20233dequivariantdiffusiontargetaware} & 0.244 & -0.344 & 0.480 & 4.738 & 35.08 & 28.07 & 21.41 & 65.19 & 53.17 & 51.28 \\
 & DrugFlow \cite{schneuing2025multidomain} & 0.252 & -0.338 & 0.553 & 3.43 & 69.97 & 65.80 & 61.41  & 75.86 & 64.84 & \textbf{78.45} \\
 & Pocket2Mol \cite{pmlr-v162-peng22b} & \textbf{0.297} & \underline{-0.425} &0.573 & 3.198 & 51.83 & 46.90 & 41.22 & \textbf{83.46} & 65.95 & 72.02 \\
\midrule
\multirow{3}{*}{\shortstack{LLM-based\\Models}} 
 & TamGen \cite{Wu2024TamGenDD} & 0.089 & -0.003 & 0.460 & 7.528 & 81.02$^\dagger$ & 37.08 & 66.50 & 37.08 & 57.75 & 5.64 \\
 & ELILLM-Diff \cite{hu2026empoweringllmsstructurebaseddrug} & 0.085 & -0.002 & 0.487 & 8.567 & 31.78 & 24.50 & 18.53 & 22.62 & 71.31$^\dagger$ & 3.60 \\
 & ELILLM-Rand \cite{hu2026empoweringllmsstructurebaseddrug} & 0.091 & -0.003 & 0.460 & 8.113 & 45.66 & 37.06 & 30.67 & 27.97 & 65.39 & 4.14 \\
\midrule
\multirow{7}{*}{\shortstack{LiFT\\Variants}} 
 & Ligand-Ref Embedding & 0.263$^\dagger$ & -0.352 & 0.480 & 3.660 & 57.88 & 54.07 & 50.54 & 59.69 & 47.14 & 71.74 \\
 & \textbf{LiFT (QED-Reference)} & 0.242 & -0.329 & 0.521 & 3.694 & 63.90 & 59.06 & 54.20 & 60.20 & 48.62 & 67.17 \\
 & \textbf{LiFT (Vina-Reference)} & 0.243 & -0.337 & 0.515 & 3.757 & 62.55 & 57.73 & 53.28 & 61.37 & 49.90 & 68.34 \\
 & \textbf{LiFT (Bal-Reference)} & 0.247 & -0.335 & 0.533 & 3.640 & 64.65 & 59.96 & 54.72 & 63.52 & 51.25 & 68.31 \\
 & \textbf{LiFT (QED-No-Reference)} & 0.243 & -0.334 & \underline{0.744} & 2.724$^\dagger$ & 80.43 & 74.88$^\dagger$ & 69.09$^\dagger$ & 83.19$^\dagger$ & 71.10 & 72.21$^\dagger$ \\
 & \textbf{LiFT (Vina-No-Reference)} & 0.255 & -0.373 & 0.732$^\dagger$ & \underline{2.662} & \textbf{83.08} & \textbf{78.86} & \textbf{74.07} & \underline{83.36} & \textbf{73.81} & \underline{73.56} \\
 & \textbf{LiFT (Bal-No-Reference)} & 0.241 & -0.341 & \textbf{0.757} & \textbf{2.659} & \underline{81.92} & \underline{77.76} & \underline{73.84} & 81.27 & \underline{71.78} & 70.73 \\
\bottomrule
\end{tabular}
}
\end{table*}

\subsection{Comprehensive Distribution Matching and Generalization}
\label{sec:Comprehensive Distribution}
To evaluate distribution matching, we compare generated molecules against the empirical ligand distribution estimated from the 100K CrossDocked2020 training complexes. We report reference-free and reference-guided variants: \textit{Ligand-Ref Embedding} directly encodes the available reference SMILES, whereas \textit{Norm No-Ref} and \textit{Norm Ref} use distribution-oriented LLM prompting without and with a ligand reference. We also include two representative property-oriented variants, \textit{Balanced Ref} and \textit{Vina No-Ref}, to examine reference-guided scaffold hopping and \textit{de novo} design without additional 3D generator fine-tuning.

Although LiFT is not specifically optimized for distribution matching, it achieves competitive WD performance across key metrics. Notably, No-Reference variants perform strongly in binding efficiency and topological ring distributions, surpassing the Ligand-Ref variant on several metrics. Their higher WD on QED and SA reflects a distributional shift rather than uniform degradation: generated molecules show improved pharmacological profiles relative to training-set ligands, which naturally moves them away from the empirical reference distribution, as indicated in Table~\ref{tab:optimization_results}.

\subsection{Property-Oriented Generation without Additional Generator Fine-Tuning}
\label{multiobjective}

Moving beyond distribution matching, we evaluate property-oriented generation in Table~\ref{tab:optimization_results}.
\textit{No-Reference} variants generally outperform their \textit{Reference} counterparts on core medicinal-chemistry and topological metrics, suggesting that strict anchoring to empirical ligands can constrain language-derived priors, whereas reference-free prompting allows exploration of alternative valid chemical regions.
Accordingly, \textit{No-Reference} configurations improve QED up to 0.757 and SA down to 2.659, while maintaining strong ring-frequency alignment ($78.86\%$ for rings $>10$), filter compliance (RDKit/REOS $>71\%$), and competitive PoseBusters validity (up to $73.56\%$).
Detailed subindicators are provided in Appendix~\ref{appendix:viz}.

Task-specific prompts further induce directional property shifts: Vina-oriented instructions bias generation toward higher binding affinity, whereas QED-centric instructions improve drug-likeness.
Rather than deterministic optimization, this semantic conditioning acts as a trend-guided prior, with text specifications corresponding to downstream empirical scores as analyzed in Appendix~\ref{app:alignment_suite}.
Overall, LiFT improves medicinal-chemistry profiles and filter compliance while preserving competitive binding and structural-validity indicators, suggesting a practical trade-off between language-derived property steering and 3D pocket compatibility.

\subsection{Ablation Studies and Diagnostic Analyses}
\label{sec:ablation_mechanics}

\textbf{Structural Module Ablation.}
Table~\ref{tab:ablation_comparison} evaluates the main conditioning modules.
The Ligand-Ref Embedding setting serves as a reference-conditioned diagnostic baseline for the downstream generator.
Removing zero-initialization slightly weakens binding, QED, and filter compliance, suggesting that zero-initialized modulation helps inject SMILES-derived priors without destabilizing the flow backbone.
Removing SCDR causes larger drops in QED, RDKit, and REOS, indicating that state-aware routing mainly benefits the drug-likeness and filter compliance of semantically steered generation.

\textbf{Cross-LLM Robustness and Selection.}
Benchmarking GPT-4o, Claude-4-Sonnet, and DeepSeek-V3 shows relatively small performance variation, suggesting that LiFT is not tied to a single language engine.
DeepSeek-V3 achieves the strongest filter compliance, while GPT-4o provides the best QED and a stable property-oriented profile; we therefore use GPT-4o as the default backbone.

\begin{table}[h]
    \centering
    \caption{Ablation analysis of core components and cross-LLM inference performance.}
    \label{tab:ablation_comparison}
    \resizebox{\columnwidth}{!}{
        \begin{tabular}{l c c c c c c}
            \toprule
            \textbf{Model Variant} & \textbf{Gnina $\uparrow$} & \textbf{Vina $\downarrow$} & \textbf{QED $\uparrow$} & \textbf{SA $\downarrow$} & \textbf{RDKit\% $\uparrow$} & \textbf{REOS\% $\uparrow$} \\
            \midrule
            \multicolumn{7}{l}{\textit{Structural/Component Ablation}} \\
            \midrule
            LiFT (Ligand-Ref Embedding) & \textbf{0.263} & \textbf{-0.352} & 0.480 & 3.660 & 59.69 & 47.14 \\
            w/o Zero-Init & 0.248
            & -0.339 & 0.470 & 3.660 & 58.00 & 47.00 \\
            w/o SCDR (naive semantic injection) & 0.245 & -0.335 & 0.464 & 3.591 & 57.28 & 44.62 \\
            \midrule
            \multicolumn{7}{l}{\textit{Generalization Across LLM Backbones}} \\
            \midrule
            LiFT (GPT-4o) & 0.241 & -0.341 & \textbf{0.757} & \textbf{2.659} & \underline{81.27} & 71.78 \\
            LiFT (Claude-4-Sonnet) & 0.249 & -0.330 & 0.700 & 2.777 & 78.20 & \underline{73.98} \\
            LiFT (DeepSeek-V3) & \underline{0.258} & \underline{-0.346} & \underline{0.707} & \underline{2.660} & \textbf{83.44} & \textbf{75.03} \\
            \bottomrule
        \end{tabular}
    }
\end{table}

\textbf{Language-Condition Diagnostics.}
Beyond architectural ablations, Appendix~\ref{app:condition_analysis} and Appendix~\ref{app:alignment_suite} examine LLM-derived molecular conditions from complementary perspectives.
Appendix~\ref{app:condition_analysis} compares random, retrieved, direct-prompted, ablated, and full Sense-Evolve-Assemble SMILES conditions under the same 3D generator, testing whether structured language conditions provide more balanced guidance than arbitrary or retrieval-based priors.
Appendix~\ref{app:alignment_suite} further analyzes whether these conditions retain trend-level semantic effects after cross-modal projection: Appendix~\ref{app:pearson_invariance} measures property-level correspondence between SMILES conditions and final molecules, Appendix~\ref{app:tsne_manifold} compares generated and empirical ligand SMILES in ECFP4 space, and Appendix~\ref{app:scdr_architecture} profiles SCDR gate behavior across training, ODE trajectories, and latent channels.

\section{Conclusion}
\label{sec:conclusion}

In this study, we introduce \textbf{LiFT}, a language-informed cross-modal framework built on flow matching for trend-guided 3D molecular generation.
Our evaluations show (i) strict ligand-reference anchoring can constrain exploration, whereas reference-free \textit{de novo} prompting enables exploration of valid chemical regions; and (ii) property-oriented prompting can improve medicinal chemistry profiles while maintaining competitive binding and structural validity.
These results demonstrate the potential of language-derived chemical priors as semantic conditions for pocket-conditioned 3D generation, enabling trend-level guidance without geometric-backbone fine-tuning.

\section*{Limitations}

While LiFT provides trend-guided language conditioning for SBDD, challenges in cross-modal generation remain. Natural language captures useful chemical and pharmacological preferences but lacks the geometric granularity for atomic-level control. Accordingly, LiFT provides trend-level guidance rather than exact property or structural control. Since LiFT relies on LLM-derived symbolic priors, its performance may vary with the chemical competence and reliability of the underlying LLM, especially where text-based chemical knowledge is sparse or biased. Future work may explore intermediate representations that better bridge language-level chemical intent and evolving 3D states, as well as richer modalities such as cryo-EM density maps or quantum interaction graphs.

Current SBDD evaluation lacks consensus on balancing distributional fidelity, such as Wasserstein Distance, with absolute property-oriented improvement. Validation on the static CrossDocked2020 benchmark enables controlled comparison but does not establish generalization across broader pocket distributions or capture real biological dynamics. Extending LiFT to additional targets, datasets, and wet-lab feedback remains important.


\bibliography{custom}

\appendix

\section*{Appendix Overview}

The supplementary material is organized into eight complementary parts that further analyze the role of language-derived molecular conditions, provide reproducibility details, extend the experimental diagnostics beyond the main paper, and clarify the methodological positioning of LiFT.

\begin{itemize}

    \item \textbf{Appendix~\ref{app:condition_analysis}: Extended Analysis of Language-Derived Molecular Conditions.}
    This appendix analyzes the intermediate SMILES conditions generated by the language agent before geometric generation.
    It includes condition-source ablations, 1D semantic evaluations, downstream 3D prior effects, and analyses of how structured language-derived conditions influence molecular generation under a fixed 3D backbone.

    \item \textbf{Appendix~\ref{app:alignment_suite}: Semantic Preservation and Diagnostic Analyses.}
    This appendix examines whether language-derived molecular conditions retain measurable trend-level effects after cross-modal projection into 3D generation.
    It includes property-level correspondence analysis, ECFP4 manifold comparisons, and SCDR routing diagnostics across training and ODE trajectories.

    \item \textbf{Appendix~\ref{appendix:reproducibility}: Experimental Details and Reproducibility.}
    This appendix reports datasets, implementation settings, hyperparameters, evaluation protocols, hardware configurations, runtime and efficiency, pocket-level uncertainty estimates, absolute docking scores, and additional reproducibility details.

    \item \textbf{Appendix~\ref{appendix:prompts}: Controlled SMILES-Condition Protocols and Prompting Templates.}
    This appendix describes the controlled prompting interfaces, information-access boundaries, task-specific steering directives, and structured generation templates used by the language agent.

    \item \textbf{Appendix~\ref{appendix:viz}: Rule-Level Structural, MedChem, and Pose Validity Diagnostics.}
    This appendix decomposes aggregate RDKit, REOS, and PoseBusters metrics into fine-grained rule-level analyses and geometric validity sub-checks.

    \item \textbf{Appendix~\ref{sec:appendix_case_study}: Qualitative Case Studies on Representative Targets.}
    This appendix provides qualitative analyses of generated molecules under different conditioning settings, including comparisons of structural validity, medicinal chemistry properties, and pocket compatibility.

    \item \textbf{Appendix~\ref{appendix:math}: Detailed Mathematical Formulations.}
    This appendix contains extended derivations and theoretical details for the proposed flow-matching and conditioning framework.

    \item \textbf{Appendix~\ref{appendix:related_comparison}: Extended Comparison with Related Work.}
    This appendix provides a systematic methodological comparison between LiFT and representative language-based, 1D--3D, and native pocket-conditioned SBDD methods, together with a focused comparison with ELILLM.

\end{itemize}

\section{Extended Analysis of Language-Derived Molecular Conditions}
\label{app:condition_analysis}

This appendix provides an extended analysis of the intermediate 1D SMILES conditions used by LiFT. 
The main experiments in the paper evaluate the final 3D generated molecules, while this appendix focuses on the symbolic molecular conditions before they are encoded by SMI-TED and injected into the 3D flow model. 
The purpose of this analysis is to isolate whether the proposed language-derived conditions are chemically meaningful and structurally diverse, rather than merely acting as arbitrary SMILES priors.

We evaluate six condition sources under the same reference-free balanced generation setting. 
All variants are evaluated on the same target-pocket set and produce SMILES conditions that are subsequently used by the same downstream 3D generation pipeline. 
The only difference among these variants is how the intermediate SMILES condition is obtained.

\begin{itemize}
    \item \textbf{C0-random}: valid SMILES are randomly sampled from the training ligand pool. 
    This setting controls for whether injecting any valid molecular string is sufficient to provide useful chemical priors.

    \item \textbf{C1-retrieved}: SMILES are selected from the training ligand pool through a non-LLM retrieval procedure. 
    This setting represents a stronger non-generative molecular-prior baseline than random sampling, while avoiding direct access to test ligand SMILES.

    \item \textbf{C2-direct}: the LLM directly generates SMILES from the pocket description and task preference, without the structured Sense-Evolve-Assemble procedure. 
    This setting controls for the effect of ordinary direct prompting.

    \item \textbf{C3-w/o-Sense}: the language agent generates SMILES without the explicit pocket-sensing stage. 
    This removes the structured extraction of local pocket features, such as geometric shape, charge state, and residue anchors.

    \item \textbf{C4-w/o-PoT}: the agent receives the pocket information but does not use Pocket-of-Thought reasoning before SMILES generation. 
    This setting isolates the role of structured intermediate reasoning.

    \item \textbf{C5-full}: the full Sense-Evolve-Assemble agent used in LiFT. 
    This setting combines pocket sensing, Pocket-of-Thought reasoning, and task-aware molecular assembly to produce the final SMILES condition.
\end{itemize}

This comparison separates three sources of signal: generic valid molecular priors (\textbf{C0-random}), non-LLM retrieved molecular priors (\textbf{C1-retrieved}), and LLM-derived molecular conditions with different levels of structured reasoning (\textbf{C2--C5}). 
Therefore, the analysis directly tests whether the full language agent provides more chemically structured 1D conditions than simpler alternatives.

\begin{table*}[t]
\centering
\small
\setlength{\tabcolsep}{5.2pt}
\renewcommand{\arraystretch}{1.08}
\caption{
1D molecular condition quality under different SMILES condition sources.
All variants are evaluated before downstream 3D generation. 
C0 samples random valid SMILES from the training ligand pool; C1 retrieves training-set SMILES; C2 uses direct LLM prompting; C3 removes explicit pocket-sensing information; C4 removes Pocket-of-Thought reasoning; and C5 is the full Sense-Evolve-Assemble agent.
The full agent preserves perfect validity, achieves the highest uniqueness, maintains high scaffold diversity, and produces chemically plausible property profiles without collapsing to retrieved training-set scaffolds.
Boldface is omitted because this table is intended to compare overall trade-offs among validity, diversity, novelty, and medicinal-chemistry plausibility rather than single-metric dominance.
}
\begin{tabular}{lcccccccc}
\toprule
\textbf{Condition Source} 
& \textbf{Validity} 
& \textbf{Unique} 
& \textbf{Scaf. Unique} 
& \textbf{QED} 
& \textbf{SA} 
& \textbf{LogP} 
& \textbf{Rings} 
& \textbf{Novel Scaf.} \\
\midrule
C0 Random 
& 1.000 
& 0.987 
& 0.777 
& 0.543 
& 3.241 
& 2.101 
& 2.877 
& 0.000 \\
C1 Retrieved 
& 1.000 
& 0.767 
& 0.000 
& 0.673 
& 2.512 
& 3.647 
& 1.977 
& 0.000 \\
C2 Direct 
& 1.000 
& 0.947 
& 0.624 
& 0.801
& 2.048
& 2.583 
& 1.466 
& 0.641 \\
C3 w/o Sense 
& 1.000 
& 0.987 
& 0.832
& 0.777 
& 2.151 
& 3.059 
& 2.096 
& 0.857 \\
C4 w/o PoT 
& 1.000 
& 0.963 
& 0.786 
& 0.754 
& 2.129 
& 2.938 
& 2.390 
& 0.814 \\
C5 Full 
& 1.000 
& 0.993 
& 0.824 
& 0.778 
& 2.239 
& 2.529 
& 2.672
& 0.750 \\
\bottomrule
\end{tabular}
\label{tab:condition_source_1d}
\end{table*}

\subsection{1D Semantic Condition Analysis}
\label{app:condition_1d}

We first evaluate the intermediate SMILES conditions before they are injected into the 3D generator.
This analysis isolates whether the language-derived molecular conditions are chemically meaningful in their own right, rather than merely serving as arbitrary textual inputs to the downstream flow model.
All variants are evaluated on the same target set and differ only in how the 1D SMILES condition is obtained.

Using the six condition sources defined above, we first evaluate the resulting SMILES before they are injected into the 3D generator.
Table~\ref{tab:condition_source_1d} reports representative 1D condition-quality metrics.
We focus on validity, uniqueness, scaffold uniqueness, QED, SA, LogP, ring count, and scaffold novelty with respect to the training set.
These metrics cover basic chemical validity, molecular diversity, medicinal-chemistry plausibility, and whether the generated conditions collapse to training-set scaffolds.

Several observations emerge.
First, validity is saturated across all variants, indicating that the comparison is not driven by invalid SMILES filtering.
Second, the retrieved baseline has zero scaffold novelty by construction and substantially lower scaffold diversity than generated conditions, showing that retrieval provides chemically valid but non-novel molecular priors.
Third, direct prompting achieves strong QED and SA, but it produces lower scaffold uniqueness and simpler ring profiles, suggesting that direct LLM generation tends to optimize local drug-likeness without maintaining the same level of topological diversity.
Finally, the full agent achieves the highest uniqueness, high scaffold uniqueness, strong scaffold novelty, and a balanced property profile.
This suggests that the Sense-Evolve-Assemble design provides useful 1D molecular conditions beyond random, retrieved, or directly prompted SMILES.

\begin{table*}[t]
\centering
\small
\setlength{\tabcolsep}{4.0pt}
\renewcommand{\arraystretch}{1.08}
\caption{
Downstream 3D generation results under different 1D SMILES condition sources.
All variants use the same SMI-TED encoder, 3D flow generator, sampling configuration, and evaluation pipeline; only the source of the intermediate SMILES condition changes.
C0 samples random valid SMILES from the training ligand pool; C1 uses retrieved training-set SMILES; C2 directly prompts the LLM to generate SMILES; C3 removes explicit pocket-sensing information; C4 removes Pocket-of-Thought reasoning; and C5 is the full Sense-Evolve-Assemble agent.
Boldface is omitted because this table evaluates trade-offs across medicinal chemistry, binding-efficiency, topological, and structural-validity indicators rather than optimizing a single metric.
}
\resizebox{\textwidth}{!}{
\begin{tabular}{lcccccccccc}
\toprule
\textbf{Condition} 
& \textbf{QED}$\uparrow$
& \textbf{SA}$\downarrow$
& \textbf{VinaEff}$\downarrow$
& \textbf{GninaEff}$\uparrow$
& \textbf{LogP}
& \textbf{Rot.}$\uparrow$
& \textbf{Ring$>$0}$\uparrow$
& \textbf{Ring$>$10}$\uparrow$
& \textbf{Ring$>$100}$\uparrow$
& \textbf{PB-Int.}$\uparrow$ \\
\midrule
C0 Random 
& 0.579 & 3.699 & -0.302 & 0.233 & 1.766 & 4.755 & 0.629 & 0.583 & 0.512 & 0.794 \\
C1 Retrieved 
& 0.693 & 2.855 & -0.318 & 0.225 & 3.159 & 3.931 & 0.837 & 0.824 & 0.797 & 0.856 \\
C2 Direct 
& 0.760 & 2.683 & -0.345 & 0.236 & 2.547 & 2.925 & 0.777 & 0.713 & 0.666 & 0.900 \\
C3 w/o Sense 
& 0.749 & 2.531 & -0.343 & 0.238 & 3.117 & 2.584 & 0.766 & 0.706 & 0.656 & 0.881 \\
C4 w/o PoT 
& 0.736 & 2.732 & -0.329 & 0.234 & 2.851 & 3.110 & 0.748 & 0.680 & 0.615 & 0.900 \\
C5 Full 
& 0.757 & 2.659 & -0.331 & 0.241 & 2.347 & 3.716 & 0.819 & 0.778 & 0.738 & 0.902 \\
\bottomrule
\end{tabular}
}
\label{tab:condition_source_3d}
\end{table*}

\subsection{Downstream Effect of 1D Condition Sources}
\label{app:condition_source_3d}

We next examine whether the differences observed at the 1D condition level translate into downstream 3D molecular generation behavior. For this analysis, all condition sources are passed through the same frozen SMI-TED encoder and the same trained 3D flow generator. The sampling configuration and evaluation pipeline are also fixed, so the only changing factor is the source of the intermediate SMILES condition.

Table~\ref{tab:condition_source_3d} reports downstream 3D generation results under the six condition sources defined in Appendix~\ref{app:condition_1d}. Random SMILES conditions lead to the weakest overall performance, with lower QED, worse SA, weaker binding-efficiency scores, and lower PoseBusters interaction-energy validity. This confirms that injecting an arbitrary valid molecular prior is insufficient for effective 3D generation. Retrieved SMILES provide stronger ring-frequency statistics, which is expected because retrieved molecules are drawn from the empirical training ligand pool. However, this retrieval-based prior does not yield the best binding-related scores and also shows a relatively high LogP, suggesting that simply reusing training-set molecular priors does not provide the most effective target-aware guidance.

Direct LLM prompting achieves the highest QED and the best Vina efficiency, indicating that an LLM can generate useful drug-like conditions even without structured reasoning. However, its ring-frequency metrics are consistently lower than those of the full agent, suggesting that direct prompting tends to favor local medicinal-chemistry optimization while providing less balanced topological guidance. Removing the Sense phase or Pocket-of-Thought reasoning also leads to weaker overall trade-offs. The w/o Sense variant obtains the lowest SA score, but it has reduced ring-frequency coverage and lower PoseBusters interaction-energy validity than the full model. Similarly, w/o PoT weakens ring statistics and does not improve binding-efficiency metrics.

The full Sense-Evolve-Assemble agent provides the most balanced downstream behavior. Although it is not the best on every single metric, it achieves the best Gnina efficiency and the highest PoseBusters interaction-energy validity, while maintaining strong QED, SA, and ring-frequency statistics. Compared with Direct LLM and w/o PoT, the full agent better preserves topological richness after 3D generation. Compared with Retrieved SMILES, it avoids collapsing to empirical training-set priors while producing stronger binding-related and structural-validity indicators. These results support the view that structured language-derived SMILES conditions act as informative soft priors for 3D generation, rather than arbitrary molecular strings or simple retrieval anchors.

Together with the 1D condition-quality analysis in Appendix~\ref{app:condition_1d}, these results indicate that the proposed language agent improves both the intermediate symbolic conditions and their downstream effect after cross-modal projection into 3D molecular generation.

\section{Semantic Preservation and Diagnostic Analyses}
\label{app:alignment_suite}

This appendix provides additional diagnostic analyses for the language-conditioned generation pipeline. Rather than treating the LLM-generated SMILES as a black-box textual input, we examine whether these 1D conditions remain chemically meaningful at different stages of the framework. The goal is not to prove exact preservation or one-to-one recovery of ground-truth ligands. Instead, we ask whether the agent-generated SMILES and their downstream 3D generations preserve measurable trend-level and distribution-level relationships with empirical molecular data.

We study this question from three complementary perspectives. Appendix~\ref{app:pearson_invariance} evaluates whether property tendencies encoded in the agent-generated SMILES are reflected in the calculated properties of the final generated 3D molecules. Appendix~\ref{app:tsne_manifold} analyzes the topo-chemical distribution of the agent-generated SMILES themselves by comparing their ECFP4 fingerprint embeddings with the 100 held-out ligand SMILES from the evaluated pockets. Appendix~\ref{app:scdr_architecture} further profiles the learned behavior of the State Calibration Gate, Scalar Update Gate, and Vector Update Gate across training, ODE inference, and latent-channel distributions. Together, these analyses provide diagnostic evidence that the language-derived conditions are not arbitrary text strings, but chemically structured inputs that remain informative before and during 3D generation.

\begin{figure*}[htbp]
    \centering
    \includegraphics[width=0.98\linewidth]{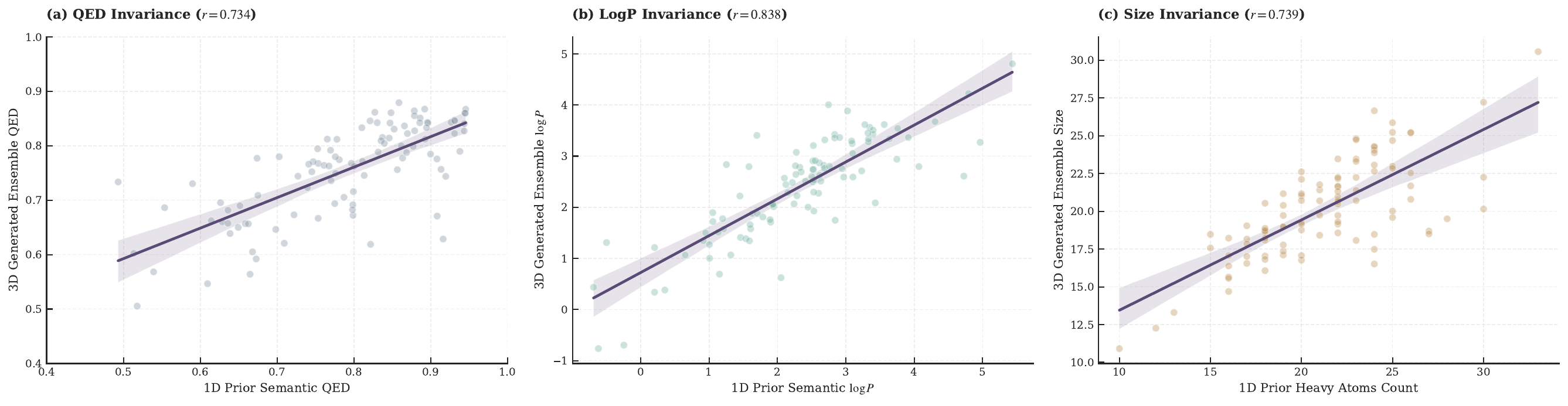}
    \caption{\textbf{Property-level trend correspondence under the \textit{LiFT (Balanced No-Ref)} setting.} 
    The panels show ensemble-level correlations between properties derived from the agent-generated SMILES conditions and calculated properties of the final generated molecules across 100 test pockets:
    (a) QED ($r=0.734$),
    (b) $\log P$ ($r=0.838$), and
    (c) molecular size measured by heavy atom count ($r=0.739$).
    All reported Pearson correlations are statistically significant ($p < 10^{-17}$).
    These results indicate that the SMILES-derived conditions preserve measurable property-level trends after cross-modal projection and 3D generation, without implying exact molecule-level regression.}
    \label{fig:fidelity_scatter}
\end{figure*}

\begin{table}[t]
\centering
\small
\setlength{\tabcolsep}{5.5pt}
\renewcommand{\arraystretch}{1.08}
\caption{
Cross-modal trend correspondence between text-derived 1D conditioning targets and the final generated 3D molecular ensembles under the \textit{LiFT (Balanced No-Ref)} setting.
Pearson correlations are computed at the pocket-ensemble level over 100 target pockets.
}
\begin{tabular}{lcc}
\toprule
\textbf{Property} & \textbf{Pearson $r$} & \textbf{$p$-value} \\
\midrule
QED & 0.7344 & $3.457{\times}10^{-18}$ \\
LogP & 0.8379 & $1.646{\times}10^{-27}$ \\
Heavy atoms & 0.7392 & $1.613{\times}10^{-18}$ \\
Rotatable bonds & 0.7619 & $3.427{\times}10^{-20}$ \\
6-ring count & 0.7578 & $7.066{\times}10^{-20}$ \\
\bottomrule
\end{tabular}
\label{tab:cross_modal_correlation}
\end{table}

\subsection{Property-Level Trend Correspondence under Cross-Modal Projection}
\label{app:pearson_invariance}

To examine whether text-derived conditions remain reflected in the generated molecules, we measure the correlation between agent-generated SMILES-derived conditions and calculated molecular properties after 3D generation. This analysis is conducted at the ensemble level rather than the single-molecule level. For each target pocket, we average the properties of its generated ligands, which reduces sampling noise and focuses on macroscopic generation trends.

Under the unconstrained \textit{LiFT (Balanced No-Ref)} setting, we generated more than 6,000 ligands for 100 target pockets from CrossDocked2020. For each pocket, we computed the average property value over its generated ligand ensemble and compared it with the corresponding text-derived conditioning target. As shown in Figure~\ref{fig:fidelity_scatter} and Table~\ref{tab:cross_modal_correlation}, the generated ensembles show consistently positive correlations across multiple physicochemical and topological dimensions. All reported correlations are statistically significant, suggesting that language-derived conditions preserve measurable trend-level signals after cross-modal projection into 3D molecular generation.

These results should be interpreted as evidence of trend-level correspondence rather than exact property control. The generated molecules are not expected to match the conditioning values point by point, since the final structures are also constrained by pocket geometry, flow dynamics, and chemical validity. Nevertheless, the observed correlations suggest that the language-derived conditions remain informative after projection into the 3D generation process. This supports the use of text-derived priors as soft guidance signals for property-oriented generation, while leaving the precise causal role of each architectural component to the ablation and diagnostic analyses.

\subsection{Topo-Chemical Distribution of Agent-Generated SMILES}
\label{app:tsne_manifold}

While Appendix~\ref{app:pearson_invariance} examines whether SMILES-derived property tendencies are reflected after 3D generation, we further evaluate the agent-generated SMILES before they are injected into the geometric generator. This analysis asks whether the 1D molecules proposed by the language agent occupy topo-chemical regions comparable to the empirical ligand SMILES from the test set.

For each molecule, we compute 2,048-bit Morgan fingerprints (ECFP4, radius $=2$) to represent local substructural patterns. We then project these high-dimensional fingerprints into two dimensions using $t$-Distributed Stochastic Neighbor Embedding ($t$-SNE). The comparison is performed between the agent-generated SMILES and the 100 ground-truth ligand SMILES corresponding to the test pockets. Since $t$-SNE is stochastic and depends on the full input matrix of each run, absolute coordinates should not be compared across panels. We therefore use the visualization only as a qualitative diagnostic of relative overlap, separation, and neighborhood structure within each panel.

\begin{figure*}[htbp]
    \centering
    \includegraphics[width=0.98\linewidth]{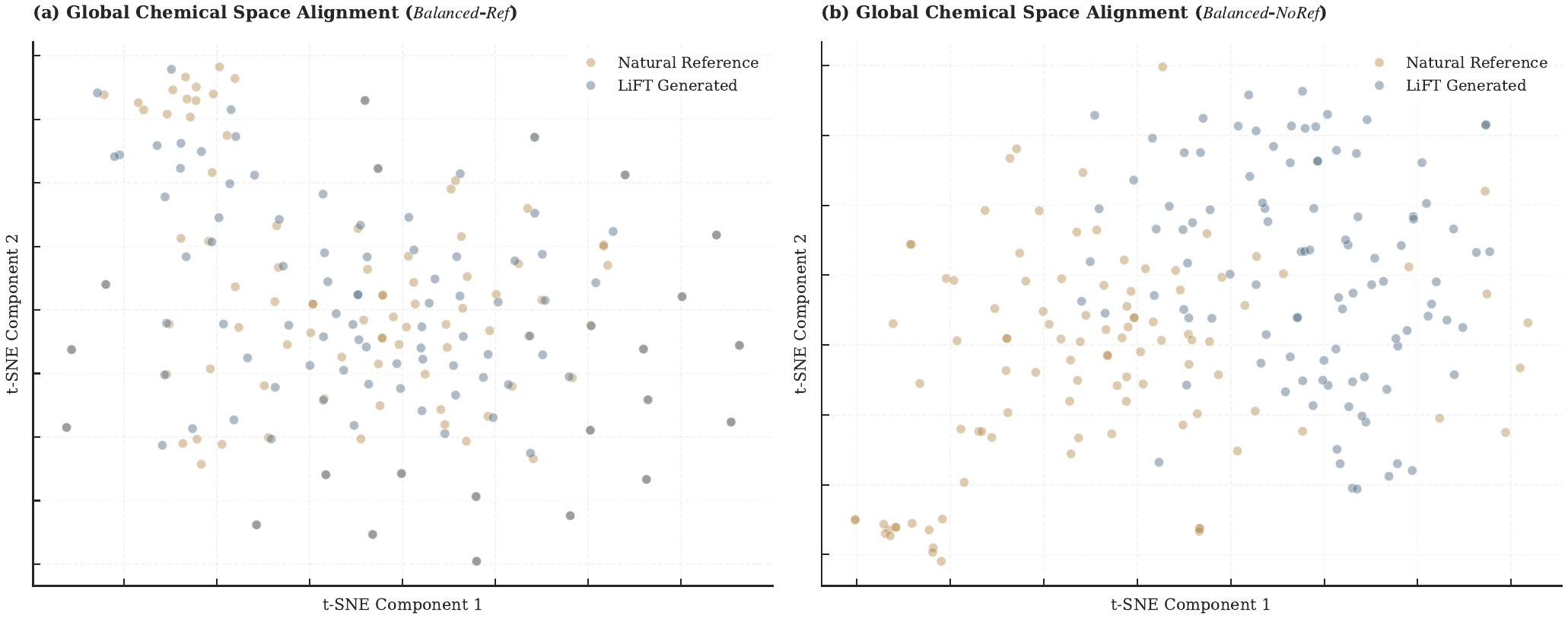}
    \caption{\textbf{Qualitative $t$-SNE visualization of topo-chemical distributions for agent-generated SMILES.}
    The plots compare ECFP4 fingerprint embeddings of SMILES produced by the language agent with the 100 ground-truth ligand SMILES from the test pockets.
    (a) Under the reference-guided \textit{LiFT (Balanced Ref)} setting, the agent-generated SMILES show broad overlap with the ground-truth ligand distribution in the projected fingerprint space.
    (b) Under the reference-free \textit{LiFT (Balanced No-Ref)} setting, the agent-generated SMILES exhibit a more visible distributional shift while remaining near the empirical SMILES distribution.
    Because $t$-SNE is a qualitative embedding method, this figure is intended to show distributional tendencies rather than precise chemical-space distances.}
    \label{fig:tsne_hologram}
\end{figure*}

As shown in Figure~\ref{fig:tsne_hologram}a, the reference-guided \textit{LiFT (Balanced Ref)} setting produces SMILES whose ECFP4 embeddings broadly overlap with the ground-truth ligand SMILES. This behavior is expected because the reference-guided setting provides an explicit molecular anchor to the agent. The observed overlap suggests that, when reference information is available, the agent tends to generate 1D molecular structures with local substructural patterns similar to those of the empirical ligands.

In the reference-free \textit{LiFT (Balanced No-Ref)} setting, shown in Figure~\ref{fig:tsne_hologram}b, the agent-generated SMILES display a clearer shift relative to the ground-truth distribution. This is also expected, since the agent does not receive a reference molecule and therefore has more freedom to propose alternative scaffolds. Importantly, the generated SMILES remain close to the empirical distribution in the projected fingerprint space, rather than forming a visually detached cluster. This suggests that the reference-free agent explores nearby topo-chemical regions while still preserving recognizable substructural similarity under the ECFP4 representation.

Overall, this analysis complements the property-level correspondence in Appendix~\ref{app:pearson_invariance}. Appendix~\ref{app:pearson_invariance} shows that the agent-generated SMILES provide property-level signals that remain visible after 3D generation. In contrast, the present analysis examines the SMILES conditions themselves and shows that their fingerprint distributions remain close to the ground-truth ligand SMILES, especially under reference-guided generation. These results do not prove exact recovery of empirical ligands, nor do they imply that $t$-SNE distances are chemically quantitative. They provide a qualitative check that the language agent produces chemically structured 1D conditions before the downstream 3D flow model uses them.

\begin{figure*}[t]
\centering
\includegraphics[width=0.98\linewidth]{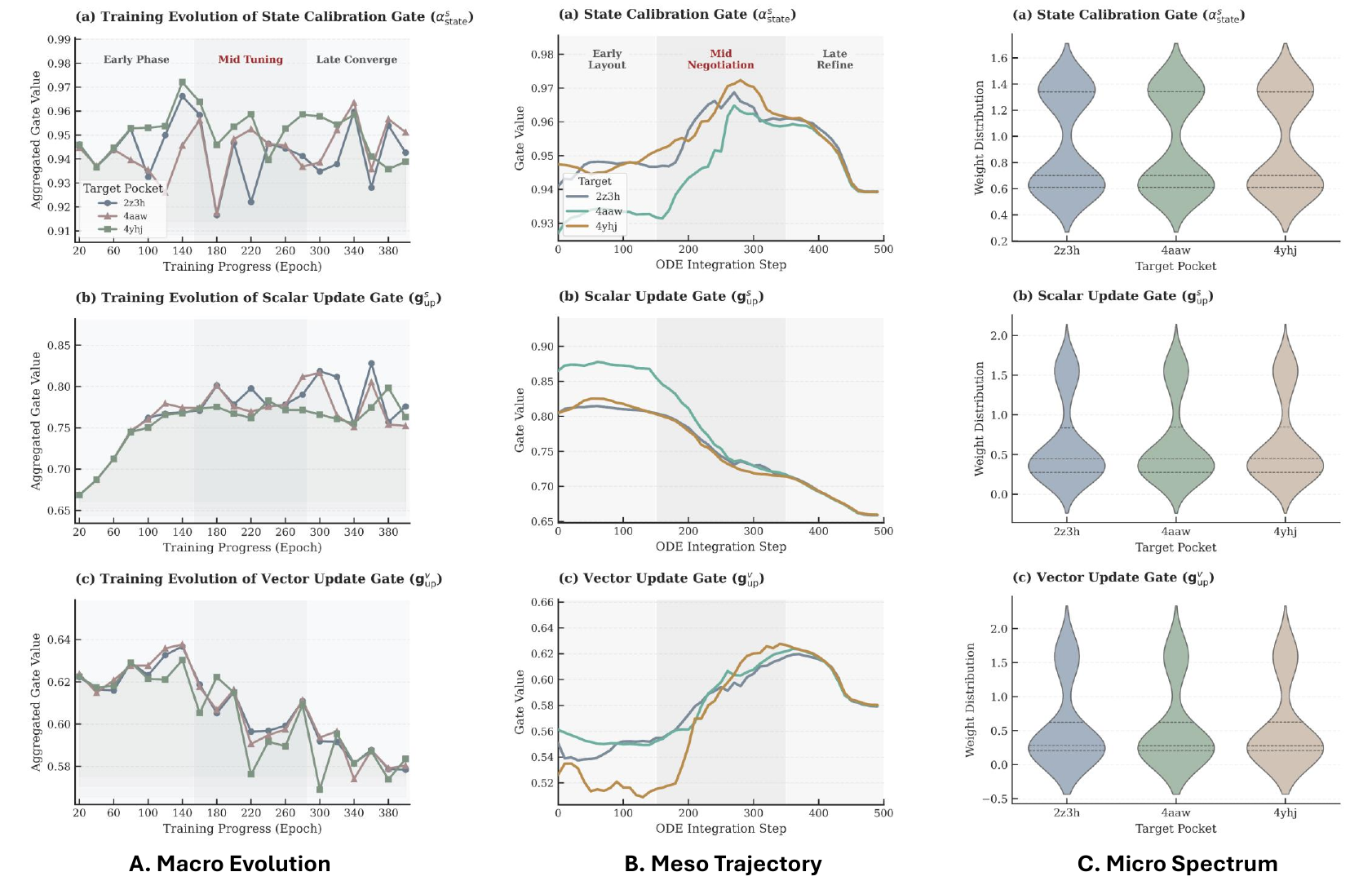}
\caption{
\textbf{Holistic spatiotemporal diagnostics of SCDR.}
This figure analyzes the learned gate behavior of SCDR rather than its architectural design.
\textbf{A. Macro evolution:} aggregated gate values across training epochs for the State Calibration Gate $\alpha_{\mathrm{state}}^{s}$, Scalar Update Gate $g_{\mathrm{up}}^{s}$, and Vector Update Gate $g_{\mathrm{up}}^{v}$.
\textbf{B. Meso trajectory:} gate values along ODE integration steps, showing different temporal patterns for scalar and vector updates and visible pocket-level variation in the middle stage.
\textbf{C. Micro spectrum:} latent-channel distributions of gate weights across target pockets, showing heterogeneous and pocket-dependent modulation patterns.
These diagnostics provide evidence that SCDR learns structured gate dynamics across training, inference, and latent-channel scales.
}
\label{fig:scdr_diagnostics}
\end{figure*}

\subsection{Holistic Spatiotemporal Diagnostics of SCDR}
\label{app:scdr_architecture}

Figure~\ref{fig:scdr_diagnostics} provides a diagnostic view of the learned behavior of the Self-Conditioned Decoupled Router (SCDR). The goal of this analysis is not to repeat the architectural design of SCDR, which has been described in Figure~3, Section~3.4, and Appendix~F. Instead, we use this figure to examine how the learned gates vary across training, ODE inference, and latent-channel distributions. We focus on three outputs: the State Calibration Gate $\alpha_{\mathrm{state}}^{s}$, the Scalar Update Gate $g_{\mathrm{up}}^{s}$, and the Vector Update Gate $g_{\mathrm{up}}^{v}$.

\paragraph{Macro-level training evolution.}
The left column of Figure~\ref{fig:scdr_diagnostics} shows the aggregated gate values across training epochs. The State Calibration Gate $\alpha_{\mathrm{state}}^{s}$ stays within a relatively high range during training, with moderate fluctuations across pockets. This suggests that the state-calibration branch remains active rather than being ignored by the model. The Scalar Update Gate $g_{\mathrm{up}}^{s}$ rises during the early stage and then stays in a more stable range, indicating that scalar-state updates are learned early and remain consistently used afterward. In contrast, the Vector Update Gate $g_{\mathrm{up}}^{v}$ shows a gradual downward trend after the early phase. This pattern is consistent with a more conservative use of vector-state updates during later training, although we do not interpret it as direct proof of a specific physical mechanism.

\paragraph{Meso-level ODE trajectory dynamics.}
The middle column profiles the same gates along the ODE integration trajectory. The State Calibration Gate $\alpha_{\mathrm{state}}^{s}$ increases during the middle stage of integration and then decreases toward the final steps. This indicates that state calibration is most active in the intermediate part of generation, when the ligand state has already moved away from the initial noise but has not yet reached the final configuration. The Scalar Update Gate $g_{\mathrm{up}}^{s}$ generally decreases along the trajectory, suggesting that scalar updates are stronger at earlier stages and gradually relaxed as integration proceeds. By comparison, the Vector Update Gate $g_{\mathrm{up}}^{v}$ increases in the middle stage before declining near the end. This shows that scalar and vector updates follow different temporal patterns during inference. Across the three target pockets, the curves differ most visibly in the middle stage but become closer near the terminal steps, suggesting pocket-dependent intermediate behavior with relatively similar final regimes.

\paragraph{Micro-level latent-channel spectra.}
The right column shows the distribution of gate weights across latent channels for different target pockets. The distributions are not uniform or single-peaked. For $\alpha_{\mathrm{state}}^{s}$, the violin plots show separated high- and mid-value regions, indicating that state calibration is unevenly distributed across channels. The Scalar Update Gate $g_{\mathrm{up}}^{s}$ and Vector Update Gate $g_{\mathrm{up}}^{v}$ also show broad and partly multi-modal distributions, with visible differences among pockets. These observations suggest that SCDR does not apply the same update strength to all latent channels. Instead, different channels receive different levels of modulation, and the resulting distributions vary with the target pocket.

\paragraph{Summary.}
Overall, Figure~\ref{fig:scdr_diagnostics} shows that SCDR has non-trivial learned gate behavior across three scales. During training, the three gates exhibit different evolution patterns. During ODE inference, scalar and vector updates follow different temporal trends, with stronger pocket-level variation in the middle of the trajectory. At the channel level, the gate distributions show heterogeneous and pocket-dependent patterns rather than uniform collapse. These results should be viewed as diagnostic evidence rather than causal proof. They support a modest interpretation: SCDR learns structured, stage-dependent modulation signals that are compatible with the ablation results in Table~\ref{tab:ablation_comparison}. Together with the property-invariance analysis in Appendix~\ref{app:pearson_invariance} and the topo-chemical manifold analysis in Appendix~\ref{app:tsne_manifold}, this provides additional evidence that the language-derived conditions behave as useful trend-level signals in the 3D generation process.

\section{Experiment Details}
\label{appendix:reproducibility}

This section describes the dataset construction, conditioning settings, baseline replication protocol, and implementation details used in our experiments. The goal is to make the evaluation protocol explicit and auditable across distribution matching, reference-guided generation, de novo generation, and preference-alignment settings.

\subsection{Dataset Preparation and Preprocessing}
\label{appendix:details_dataset}

We use the refined \textit{CrossDocked2020} benchmark for structure-based molecular generation. Each example is derived from a protein--ligand complex and provides two complementary views: a local three-dimensional pocket--ligand geometry for training the generative dynamics, and a one-dimensional molecular representation for constructing semantic conditions. This setting matches the target-aware generation problem considered in LiFT, where ligand generation is conditioned on explicit pocket geometry while semantic information is provided through a controlled molecular prior.

\textbf{Complex curation.}
We follow the standard refined CrossDocked preprocessing protocol and construct examples from the provided pocket-level structures. Complexes with unreliable binding poses, invalid molecular structures, or failed feature construction are excluded. Ligands are restricted to the small-molecule regime relevant to SBDD, ensuring that the benchmark focuses on realistic pocket-conditioned ligand generation rather than out-of-distribution molecular sizes.

\textbf{Structural condition.}
For each retained complex, the receptor is represented by the local binding pocket defined by the benchmark preprocessing protocol. This pocket definition is used only to construct the geometric conditioning graph and should not be confused with reference-ligand information exposed to the LLM. In the no-reference setting, the LLM does not receive the ground-truth ligand identity, SMILES, docking scores, or molecular properties. The protein pocket is encoded as a compact residue-level graph with scalar residue features and geometry-aware vector features, while the ligand is represented by coordinates, atom types, and bond types for flow-matching training and sampling.

\textbf{Semantic condition.}
For semantic embedding construction, ligand strings are canonicalized with RDKit and encoded by the frozen SMI-TED encoder into a fixed-dimensional molecular semantic vector. This vector provides the one-dimensional chemical prior used by the cross-modal conditioning interface. For controlled comparisons, we construct Ligand-Ref and pure LLM embedding variants under the same processed protein--ligand split, so that the compared settings share identical pocket geometries and differ only in the origin of the semantic condition.

\textbf{Evaluation.}
After preprocessing, the training split contains approximately $10^5$ refined complexes, and a held-out subset of 100 protein pockets is used for in-distribution evaluation. For each held-out pocket, LiFT generates 100 samples using the same inference configuration across all reported settings. The resulting molecules are evaluated with a unified post-processing and metric-computation pipeline, including cheminformatics validity checks, structural filters, docking-based scores, and distributional property analyses.

\subsection{Baseline Specification and Replication Protocol}
\label{appendix:details_baselines}

We compare LiFT with representative structure-based molecular generation methods spanning four major modeling families: autoregressive construction, 3D diffusion, flow matching, and LLM-based SBDD. This benchmark suite covers the dominant design choices in target-aware ligand generation, including sequential graph growth, iterative geometric denoising, continuous-time generative transport, and language-guided molecular proposal.

\textbf{Autoregressive Models.}
Autoregressive methods formulate ligand generation as a sequence of local construction decisions conditioned on the protein pocket. \texttt{AR} \cite{luo20223dgenerativemodelstructurebased} represents early graph-based atom-by-atom generation for SBDD, while \texttt{Pocket2Mol} \cite{pmlr-v162-peng22b} extends this direction with an equivariant pocket-aware representation for predicting atom types, coordinates, and bonds during molecular growth. These methods provide a representative comparison for discrete molecular construction under structural constraints.

\textbf{3D Diffusion Models.}
Diffusion-based methods generate ligands by learning to denoise molecular states in three-dimensional space under pocket conditioning. \texttt{TargetDiff} \cite{guan20233dequivariantdiffusiontargetaware} serves as a canonical equivariant diffusion baseline for target-aware generation. \texttt{DecompDiff} \cite{guan2024decompdiffdiffusionmodelsdecomposed} incorporates decomposed ligand priors, \texttt{BindDM} \cite{huang2024bindingadaptivediffusionmodelsstructurebased} emphasizes binding-adaptive generation. Together, these baselines represent the current mainstream of 3D geometric generative modeling for SBDD.

\textbf{Flow-Matching Models.}
Flow-matching methods learn continuous transport dynamics from simple priors to pocket-conditioned ligand distributions. We include \texttt{PAFlow} \cite{zhou2025priorguided} and \texttt{DrugFlow} \cite{schneuing2025multidomain} as representative continuous-time SBDD baselines. This group provides the most direct comparison for evaluating generative models based on learned vector fields and ODE-style sampling.

\textbf{LLM-based SBDD Pipelines.}
LLM-based pipelines examine whether molecular language priors can support target-aware ligand design under structure-based evaluation. \texttt{TamGen} \cite{Wu2024TamGenDD} represents target-conditioned molecular generation with language-modeling components. \texttt{ELILLM} variants, including \texttt{ELILLM-Diff} and \texttt{ELILLM-Rand} \cite{hu2026empoweringllmsstructurebaseddrug}, are included as recent LLM-driven SBDD baselines. This group evaluates how language-guided molecular proposal pipelines compare with explicitly 3D generative models under the same benchmark.

\textbf{Replication protocol.}
We distinguish between full reproduction and unified re-evaluation. For \texttt{DrugFlow} and \texttt{TamGen}, we fully reproduce the baselines using the officially released configuration files. In both cases, we directly follow the provided YAML settings and do not modify model hyperparameters, sampling parameters, or post-processing options. \texttt{DrugFlow} is included as the closest flow-matching baseline to LiFT, while \texttt{TamGen} is re-run because its generated molecule files are not publicly released in a directly usable form.

For the remaining baselines, we use the publicly released generated molecules or official sample files corresponding to the authors' recommended evaluation setting whenever available, and evaluate them with our unified evaluator. This choice avoids introducing additional implementation variance from re-training or re-sampling multiple heterogeneous models. All molecules, including those generated by LiFT, are processed with the same validity checking, docking/rescoring, PoseBusters auditing, RDKit/REOS filtering, and metric-computation pipeline. Therefore, for these methods, the comparison should be interpreted as a unified re-evaluation of released samples rather than a full reproduction of the original training and sampling procedure.

\subsection{Detailed Metric Definitions}
\label{appendix:details_metrics}

To ensure an explicit and auditable evaluation protocol, our metrics strictly reflect the experimental multi-objective tracks (Sec. \ref{multiobjective}), evaluating both absolute generative maximization (Means/Pass Rates) and empirical distribution alignment (Wasserstein Distance). 

\textbf{3D Binding Efficiencies.} To rigorously assess target-aware affinity while penalizing physically unrealistic molecular inflation, both primary docking metrics are computed as efficiencies (normalized by the generated ligand's heavy atom count). \textit{Vina Efficiency} is derived from the raw AutoDock Vina v1.2.5 binding free energy ($\text{kcal/mol}$). Analogously, \textit{Gnina Efficiency} is computed using the GNINA framework, leveraging its 3D CNN-based cross-docking scoring function to evaluate deep empirical pose quality per heavy atom.

\textbf{1D Pharmacological \& Topological Properties.} Continuous and discrete molecular properties are computed deterministically via \texttt{RDKit}. This track evaluates \textit{QED} (Quantitative Estimate of Drug-likeness) and \textit{LogP} (partition coefficient) for pharmacological viability, alongside the \textit{SA} (Synthetic Accessibility) score to assess synthesis feasibility. Geometric complexity and spatial flexibility are systematically quantified via \textit{Rotatable Bond Counts} and \textit{Topological Ring Frequencies} (categorized into $>0$, $>10$, and $>100$ ring counts).

\textbf{Structural Validity \& MedChem Filters.} Physical realism and medicinal chemistry safety are audited through comprehensive sub-filter suites. \textit{PoseBusters} pass rates strictly verify 3D geometric constraints, covering critical sub-checks including Ring Flatness, Bond Angles, Double Bond stereochemistry, Internal Steric Clashes, and Protein-Ligand Overlap limits. Furthermore, 1D structural alerts are rigorously enforced via \textit{REOS} (rapidly filtering BMS, Glaxo, Inpharmatica, and generic PAINS rule families) and \textit{RDKit Structural Alerts} (comprehensively detecting NIH, PAINS-A/B/C, and ZINC violations) to exclude toxic or chemically reactive substructures.

\textbf{Property Distribution Divergence.} To quantify the macroscopic density shifts between the generated ensembles and the empirical target distribution across all aforementioned continuous metrics (Vina, Gnina, QED, SA, LogP, Rotatable Bonds, and Ring Frequencies), we explicitly employ the \textit{Wasserstein Distance} (WD) computed via \texttt{SciPy} as our sole divergence metric. We deliberately avoid Kullback-Leibler (KL) divergence, Jensen-Shannon (JS) divergence, and Fr\'echet ChemNet Distance (FCD). This design choice is mathematically imperative for continuous-time flow matching: because empirical properties generated during unconstrained multi-objective optimization frequently exhibit disjoint supports relative to the natural distribution, KL divergence inherently suffers from numerical instability ($\infty$), while JS divergence saturates, losing directional gradient sensitivity. Conversely, the Wasserstein metric (Earth Mover's Distance) preserves a continuous, informative cost gradient even under non-overlapping supports, perfectly corresponding with our generative velocity field trajectory alignment.

\subsection{Implementation and Hyperparameters}
\label{appendix:details_hyperparams}

LiFT is implemented in \textit{PyTorch} with PyTorch Lightning. We use a frozen SMI-TED encoder to provide the ligand-level semantic condition, and train a heterogeneous GVP-GNN as the pocket-conditioned geometric backbone. The SCDR/AdaLN controller incorporates the SMILES embedding and flow-time information into ligand-node updates.
The model is optimized with a flow-matching objective over ligand coordinates, atom types, and bond types. During inference, samples are generated by integrating the learned vector field with an explicit Euler solver. We run all experiments on a single NVIDIA A800-SXM4-80GB GPU. On this hardware, one training epoch takes less than 10 minutes. For the 100-pocket evaluation setting with 100 generated ligands per pocket (10,000 molecules in total), LiFT 3D flow sampling takes 2 h 38 min 48 s; including language-condition generation and SMI-TED encoding, the complete measured pipeline takes 2 h 57 min 20 s. Docking and offline metric computation are excluded because they are evaluation procedures shared across methods rather than part of generation. The core implementation details and hyperparameter settings are summarized in Table~\ref{tab:hyperparameters}.

\begin{table}[t]
\centering
\caption{Core implementation settings of LiFT, including the GVP-based geometric backbone, SMI-TED semantic conditioning, flow-matching training objective, and Euler-based sampling protocol.}
\label{tab:hyperparameters}
\setlength{\tabcolsep}{4.5pt}
\renewcommand{\arraystretch}{1.08}
\resizebox{\columnwidth}{!}{
\begin{tabular}{lll}
\toprule
\textbf{Component} & \textbf{Hyperparameter} & \textbf{Value} \\
\midrule
\multirow{4}{*}{\textit{Backbone}}
& Geometric network & Heterogeneous GVP-GNN \\
& GVP layers & $5$ \\
& Node / edge hidden dim. & $(128,32)$ / $(128,32)$ \\
& Self-conditioning & Enabled \\
\midrule
\multirow{4}{*}{\textit{Semantic control}}
& Semantic encoder & Frozen SMI-TED \\
& SMILES embedding dim. & $768$ \\
& Controller & SCDR/AdaLN \\
& Controller condition & SMILES embedding + flow time \\
\midrule
\multirow{5}{*}{\textit{Training}}
& Objective & Flow matching over coords., atoms, and bonds \\
& Optimizer & AdamW with AMSGrad \\
& Learning rate & $8\times10^{-4}$ \\
& Batch size & $48$ \\
& Training epochs & $500$ \\
\midrule
\multirow{4}{*}{\textit{Sampling}}
& ODE solver & Explicit Euler \\
& Training flow steps & $5000$ \\
& Inference steps (NFE) & $500$ \\
& Atom / bond priors & Marginal / uniform \\
\bottomrule
\end{tabular}}
\end{table}

\paragraph{Runtime and efficiency.}
To make the inference cost explicit, Table~\ref{tab:runtime_breakdown} reports component-wise wall-clock measurements under the same A800 setup. Language-condition generation produces three valid SMILES conditions per pocket, yielding 300 conditions over 100 pockets. The 3D stage then generates 100 ligands per pocket. SMI-TED timing is reported both with and without its one-time loading cost.

\begin{table*}[t]
\centering
\caption{Measured runtime of the LiFT pipeline on a single NVIDIA A800-SXM4-80GB GPU. The matched DrugFlow comparison uses the same 100-pocket, 10,000-molecule sampling workload.}
\label{tab:runtime_breakdown}
\setlength{\tabcolsep}{5pt}
\renewcommand{\arraystretch}{1.08}
\resizebox{\textwidth}{!}{
\begin{tabular}{llll}
\toprule
\textbf{Component / Method} & \textbf{Timed Batch} & \textbf{Total Wall-Clock Time} & \textbf{Normalized Runtime} \\
\midrule
LLM condition generation & 100 pockets; 300 conditions & 14 min 48 s & 8.88 s/pocket; 2.96 s/condition \\
SMI-TED encoding, incl. loading & 300 conditions & 224.08 s & 0.747 s/condition \\
SMI-TED core encoding & 300 conditions & 208.17 s & 0.694 s/condition \\
One-time SMI-TED loading & One initialization & 14.87 s & Not amortized across future batches \\
\textbf{LiFT 3D flow sampling} & 100 pockets $\times$ 100 ligands & \textbf{2 h 38 min 48 s} & \textbf{0.953 s/molecule; 95.28 s/pocket} \\
DrugFlow 3D flow sampling & Matched 10,000-molecule workload & 1 h 59 min 20 s & 0.716 s/molecule; 71.60 s/pocket \\
\textbf{Complete measured LiFT pipeline} & LLM + SMI-TED + 3D sampling & \textbf{2 h 57 min 20 s} & \textbf{106.40 s/pocket} \\
\bottomrule
\end{tabular}}
\end{table*}

The 3D sampling stage accounts for approximately 89.5\% of the measured LiFT runtime. Under the matched sampling setting, LiFT requires 2 h 38 min 48 s versus 1 h 59 min 20 s for DrugFlow, corresponding to a 33.1\% sampling-time overhead. This comparison isolates sampling cost rather than end-to-end system cost, since DrugFlow does not include corresponding language-generation or semantic-encoding stages.

\subsection{Pocket-Level Uncertainty and Statistical Reporting}
\label{appendix:uncertainty}

To quantify variability across targets, we report pocket-level sample standard deviation (SD), across-pocket variance, and pocket-bootstrap 95\% confidence intervals (CIs). For each metric, we first aggregate generated molecules within each pocket and then compute SD and variance across test pockets. For CIs, we resample complete pockets with replacement, retain all molecules associated with each selected pocket, and recompute the original aggregate estimator. Using pockets rather than individual molecules as the resampling unit avoids treating molecules generated for the same target as statistically independent.

These quantities characterize across-pocket heterogeneity and sampling uncertainty; they should not be interpreted as retraining-seed variance. Because the reported results use fixed trained checkpoints, multi-seed retraining variance is a distinct source of uncertainty. Table~\ref{tab:pocket_uncertainty} reports SD, variance, and 95\% CIs for the continuous metrics most central to the main comparison. SD is reported in the original metric units, whereas variance is in squared units.

\begin{table*}[t]
\centering
\caption{Pocket-level uncertainty for continuous evaluation metrics. Each entry is \textbf{SD / variance / 95\% bootstrap CI}. Confidence intervals are computed by resampling complete test pockets.}
\label{tab:pocket_uncertainty}
\setlength{\tabcolsep}{3.5pt}
\renewcommand{\arraystretch}{1.06}
\resizebox{\textwidth}{!}{
\begin{tabular}{lcccc}
\toprule
\textbf{Method} & \textbf{Gnina Eff.} & \textbf{Vina Eff.} & \textbf{QED} & \textbf{SA} \\
\midrule
AR & 0.061/0.0038/[0.283,0.307] & 0.083/0.0068/[-0.426,-0.393] & 0.119/0.0140/[0.485,0.533] & 0.895/0.8019/[4.109,4.470] \\
PAFlow & 0.040/0.0016/[0.254,0.270] & 0.092/0.0084/[-0.445,-0.414] & 0.101/0.0102/[0.471,0.511] & 0.830/0.6882/[4.720,5.058] \\
BindDM & 0.044/0.0019/[0.240,0.257] & 0.074/0.0055/[-0.376,-0.347] & 0.100/0.0101/[0.489,0.527] & 0.654/0.4281/[4.661,4.902] \\
DecompDiff & 0.037/0.0014/[0.197,0.213] & 0.069/0.0048/[-0.281,-0.252] & 0.137/0.0187/[0.425,0.483] & 0.551/0.3041/[4.407,4.649] \\
TargetDiff & 0.043/0.0018/[0.236,0.253] & 0.069/0.0047/[-0.357,-0.330] & 0.111/0.0124/[0.458,0.500] & 0.593/0.3516/[4.624,4.850] \\
DrugFlow & 0.043/0.0018/[0.243,0.260] & 0.081/0.0065/[-0.353,-0.322] & 0.135/0.0184/[0.526,0.578] & 0.701/0.4915/[3.295,3.570] \\
Pocket2Mol & 0.063/0.0039/[0.285,0.310] & 0.076/0.0058/[-0.440,-0.410] & 0.091/0.0083/[0.555,0.591] & 0.631/0.3976/[3.076,3.323] \\
TamGen & 0.008/0.0001/[0.087,0.090] & 0.013/0.0002/[-0.006,-0.001] & 0.078/0.0060/[0.445,0.476] & 0.387/0.1499/[7.450,7.602] \\
ELILLM-Diff & 0.020/0.0004/[0.081,0.089] & 0.010/0.0001/[-0.004,-0.000] & 0.105/0.0111/[0.466,0.507] & 0.660/0.4361/[8.434,8.694] \\
ELILLM-Rand & 0.023/0.0005/[0.087,0.096] & 0.012/0.0001/[-0.005,-0.001] & 0.145/0.0210/[0.431,0.488] & 0.624/0.3893/[7.988,8.234] \\
LiFT (Lig.-Ref) & 0.075/0.0056/[0.241,0.270] & 0.098/0.0096/[-0.365,-0.326] & 0.181/0.0327/[0.444,0.516] & 1.133/1.2843/[3.435,3.892] \\
LiFT (QED-Ref) & 0.058/0.0034/[0.231,0.253] & 0.092/0.0086/[-0.347,-0.311] & 0.185/0.0342/[0.483,0.558] & 1.097/1.2029/[3.464,3.922] \\
LiFT (Vina-Ref) & 0.064/0.0040/[0.231,0.255] & 0.091/0.0083/[-0.347,-0.311] & 0.182/0.0331/[0.479,0.552] & 1.075/1.1560/[3.537,3.981] \\
LiFT (Bal-Ref) & 0.063/0.0040/[0.235,0.260] & 0.096/0.0091/[-0.354,-0.317] & 0.180/0.0324/[0.496,0.569] & 1.086/1.1794/[3.416,3.876] \\
LiFT (QED-NoRef) & 0.038/0.0014/[0.236,0.250] & 0.082/0.0068/[-0.357,-0.325] & 0.094/0.0089/[0.723,0.763] & 0.662/0.4380/[2.602,2.851] \\
LiFT (Vina-NoRef) & 0.043/0.0018/[0.248,0.264] & 0.092/0.0085/[-0.363,-0.327] & 0.099/0.0098/[0.708,0.750] & 0.593/0.3511/[2.538,2.794] \\
LiFT (Bal-NoRef) & 0.037/0.0014/[0.234,0.248] & 0.101/0.0103/[-0.350,-0.311] & 0.086/0.0074/[0.740,0.774] & 0.574/0.3291/[2.552,2.774] \\
\bottomrule
\end{tabular}}
\end{table*}

For the no-reference property-oriented variants, we additionally report uncertainty for representative filter and pose-validity metrics in Table~\ref{tab:filter_uncertainty}. For percentage metrics, SD is measured in percentage points and variance in squared percentage points.

\begin{table}[t]
\centering
\caption{Pocket-level uncertainty for representative filter and pose-validity metrics. Each entry is \textbf{SD / variance / 95\% bootstrap CI}.}
\label{tab:filter_uncertainty}
\setlength{\tabcolsep}{3.5pt}
\renewcommand{\arraystretch}{1.06}
\resizebox{\columnwidth}{!}{
\begin{tabular}{lccc}
\toprule
\textbf{Variant} & \textbf{RDKit (\%)} & \textbf{REOS (\%)} & \textbf{PoseB. (\%)} \\
\midrule
QED-NoRef & 10.92/119.2/[81.11,85.12] & 16.94/287.1/[67.92,74.12] & 21.52/462.9/[68.20,76.00] \\
Vina-NoRef & 11.83/140.0/[80.83,85.74] & 18.14/329.2/[70.17,77.25] & 21.43/459.4/[69.40,77.30] \\
Bal-NoRef & 13.76/189.3/[78.35,83.99] & 19.52/381.1/[67.88,75.58] & 21.53/463.5/[66.50,74.60] \\
\bottomrule
\end{tabular}}
\end{table}

The uncertainty estimates support a cautious interpretation of small numerical differences. In particular, the QED-NoRef variant has a QED 95\% CI of [0.723, 0.763] and an SA 95\% CI of [2.602, 2.851], while confidence intervals for several docking and pose-validity comparisons overlap with strong 3D baselines. We therefore use these statistics to characterize stability and across-pocket heterogeneity rather than to claim significance from small point-estimate gaps.

\subsection{Absolute Docking Scores}
\label{appendix:raw_docking}

For direct comparability with studies reporting unnormalized docking scores, we additionally report raw Vina and Gnina scores under the same evaluation and failure-handling protocol. These absolute scores complement the heavy-atom-normalized docking-efficiency metrics and the Wasserstein-distance analysis reported in the main paper.

\begin{table}[t]
\centering
\caption{Absolute docking scores under the unified evaluation protocol. Lower raw Vina is better, while higher raw Gnina is better. Values are mean $\pm$ standard deviation.}
\label{tab:raw_docking}
\setlength{\tabcolsep}{5pt}
\renewcommand{\arraystretch}{1.05}
\resizebox{\columnwidth}{!}{
\begin{tabular}{lcc}
\toprule
\textbf{Method} & \textbf{Raw Vina $\downarrow$} & \textbf{Raw Gnina $\uparrow$} \\
\midrule
AR & $-6.756 \pm 2.462$ & $4.741 \pm 1.391$ \\
PAFlow & $-9.647 \pm 3.503$ & $5.765 \pm 1.384$ \\
BindDM & $-8.236 \pm 3.066$ & $5.593 \pm 1.403$ \\
DecompDiff & $-7.647 \pm 2.283$ & $5.849 \pm 1.072$ \\
TargetDiff & $-7.770 \pm 2.761$ & $5.475 \pm 1.388$ \\
DrugFlow & $-6.938 \pm 2.908$ & $5.118 \pm 1.302$ \\
Pocket2Mol & $-6.984 \pm 2.907$ & $4.687 \pm 1.441$ \\
TamGen & $-0.143 \pm 1.840$ & $3.880 \pm 0.506$ \\
ELILLM-Diff & $-0.071 \pm 3.446$ & $4.420 \pm 1.182$ \\
ELILLM-Rand & $-0.133 \pm 2.608$ & $4.275 \pm 1.311$ \\
LiFT (Ligand-Ref Embedding) & $-6.825 \pm 2.555$ & $4.902 \pm 1.329$ \\
LiFT (QED-Reference) & $-6.804 \pm 2.477$ & $4.910 \pm 1.289$ \\
LiFT (Vina-Reference) & $-6.782 \pm 2.447$ & $4.906 \pm 1.310$ \\
LiFT (Balanced-Reference) & $-6.797 \pm 2.542$ & $4.902 \pm 1.305$ \\
LiFT (QED-No-Reference) & $-6.748 \pm 2.160$ & $4.785 \pm 1.057$ \\
LiFT (Vina-No-Reference) & $-6.362 \pm 2.177$ & $4.645 \pm 1.009$ \\
LiFT (Balanced-No-Reference) & $-6.576 \pm 2.601$ & $4.791 \pm 1.047$ \\
\bottomrule
\end{tabular}}
\end{table}

All methods use the same docking and failure-handling protocol. Raw values near zero for some sequence-based baselines reflect unsuccessful or unscored post-hoc 3D poses under this unified pipeline rather than a different scoring function. Taken together, the reported docking results provide three complementary views: absolute Vina/Gnina scores for conventional comparability, heavy-atom-normalized efficiency for size-aware affinity evaluation, and Wasserstein Distance for distributional fidelity.

\section{Controlled SMILES-Condition Protocol}
\label{appendix:prompts}

The main text uses the \textit{Sense-Evolve-Assemble} agent to convert pocket descriptions, task directives, and optional ligand references into one-dimensional SMILES conditions for the downstream flow model. This appendix formalizes the operational interface between the language-based SMILES proposal stage and the geometric generator. It specifies the information available to the language model, the deterministic parsing and validation procedure, and the construction of the fixed SMILES-derived semantic latent used during flow sampling.

\subsection{Semantic Intervention and Information Boundary}

LiFT conditions the geometric generator through a semantic vector $\mathbf{z}_{sem}$ produced by a frozen SMI-TED encoder. The controlled variable is the source of the SMILES string used to instantiate this vector. When the semantic condition is instantiated from an LLM-generated SMILES string, the string is produced by the fixed protocol described below. In the \textit{Ligand-Ref Embedding} condition, the available reference ligand SMILES is encoded directly as a reference embedding. This condition is used as a diagnostic, reference-guided setting to assess how reference semantic information affects the downstream generator. It is therefore reported separately from reference-free \textit{de novo} variants.

The access boundary is fixed before any target is evaluated. In the reference-free \textit{de novo} setting, the LLM receives only the pocket-derived profile and a task directive. It does not receive the test ligand, reference SMILES, hidden labels, docking scores, filter outcomes, or generated 3D structures. In the reference-guided setting, the LLM additionally receives the reference ligand SMILES and coarse reference-derived size constraints. We therefore report this setting separately as a \textit{Ligand-Ref} variant. Across both settings, downstream metrics are used only for evaluation, not for selecting, revising, or ranking the SMILES condition. The corresponding prompt skeletons make this input separation explicit: Table~\ref{box:denovo_prompt} lists the reference-free fields, while Table~\ref{box:ref_prompt} shows the additional reference-guided fields.

\subsection{Operational Protocol}

The protocol follows the three stages described in the methodology. First, the \textit{Sense} stage summarizes the target pocket into a compact textual profile, including coarse geometry and physicochemical anchors from the local interaction shell. This profile is the only target-specific input in the reference-free setting. Second, the \textit{Evolve} stage applies a fixed Pocket-of-Thought (PoT) schema. The schema asks the LLM to state a compact spatial and topological plan before emitting SMILES, making the intended SMILES condition inspectable without using the intermediate plan as a reward, filter, or selection signal. Third, the \textit{Assemble} stage extracts tagged SMILES strings, applies deterministic cheminformatics verification, and passes the first valid, SMI-TED-encodable molecule to the semantic encoder.

Prompt variants are fixed at the script level. Distribution-oriented, balanced, Vina-oriented, and QED/safety-oriented directives modify only the task text supplied to the LLM. They do not change the generator weights, invoke an additional guidance model, or introduce feedback from downstream evaluation. The same parsing, validation, and semantic extraction rule is used for all prompt families and LLM backbones considered in the experiments.

\subsection{Validation and Candidate Accounting}

Candidate acceptance is based only on molecular string validity and successful SMI-TED encoding. RDKit is used as an unweighted validity check rather than as a ranking surrogate. If a generated string fails to parse, the repair routine is restricted to local SMILES syntax correction using the parser diagnostic. It does not receive docking scores, property scores, reference-similarity scores, filter results, or sampled 3D coordinates. Thus, the repair step provides syntax-level recovery only and does not introduce score-feedback optimization over evaluation metrics. The repair interface is specified in Table~\ref{box:kgd_prompt}, which exposes only the parser diagnostic, invalid SMILES string, and captured PoT context.

When multiple valid candidates are available, their chronological order is preserved and the first SMI-TED-encodable candidate instantiates $\mathbf{z}_{sem}$. If no valid encodable candidate is produced under the fixed protocol, the target is logged as a semantic-instantiation failure rather than being replaced through post-hoc search. This accounting rule prevents downstream scores from influencing which semantic condition is passed to the 3D generator.

\subsection{Semantic Feature Extraction and Claim Boundary}

After validation, the selected SMILES string is embedded by the frozen SMI-TED encoder:
\[
\mathbf{z}_{sem} = \mathrm{Encoder}_{\mathrm{SMI\text{-}TED}}(S_{smiles}).
\]
The resulting vector is constructed before any 3D geometric generation and is held fixed throughout flow sampling. The downstream flow model does not call the LLM, rerun repair, inspect candidate rationales, or revise $\mathbf{z}_{sem}$ using generated coordinates or evaluation metrics.

This protocol supports a deliberately narrow operational claim: LiFT uses language to instantiate a fixed SMILES-derived semantic condition under predefined information boundaries. It does not claim that PoT rationales are faithful explanations, that the LLM independently performs molecular optimization, or that \textit{Ligand-Ref Embedding} represents a reference-free deployment condition. Instead, the protocol separates SMILES-condition construction from 3D generation and downstream metric evaluation.

\subsection{Prompt Skeletons and Structural Schema}
\label{subsec:prompt_skeletons}

For reproducibility, the accompanying prompt tables define the input fields, required output tags, parsing schema, reference-guided additions, and repair interface used by the implementation. Runtime variable interpolations are shown in square brackets, e.g., \texttt{[Target ID]}. These templates are fixed before evaluation and are not edited per target after observing generated molecules or benchmarking scores. Specifically, Table~\ref{box:denovo_prompt} defines the reference-free \textit{de novo} template, Table~\ref{box:ref_prompt} defines the \textit{Ligand-Ref} scaffold-hopping template, and Table~\ref{box:kgd_prompt} defines the syntax-repair interface.
\subsection{Task-Specific Property Optimization Directives}
\label{subsec:property_optimization}

To evaluate controllable property steering, we construct task-specific SMILES-derived semantic conditions by varying only the directive given to the LLM. All other components are kept unchanged: the information boundary, deterministic SMILES parsing, validation procedure, candidate-accounting rule, SMI-TED semantic extraction, and the downstream 3D generator. As a result, the balanced, Vina-oriented, and QED/safety-oriented settings isolate the effect of semantic task directives under the same retraining-free generation interface, rather than introducing separate optimization pipelines. The concrete directive blocks are reported in Table~\ref{box:opt_balanced}, Table~\ref{box:opt_safety}, and Table~\ref{box:opt_affinity}.

\begin{table*}[htbp]
\centering
\small
\begin{tabular}{|p{0.95\textwidth}|}
\hline
\rowcolor[HTML]{F3F3F3} \textbf{Box 1: Blind \textit{De Novo} Generation Template (No-Reference Setting)} \\ \hline
\textbf{[System Role]} \\
\texttt{You are a premier De Novo Drug Designer. Your task is to generate novel lead-like molecules from scratch that perfectly complement the provided 3D protein pocket environment.} \\
\\
\texttt{\textbf{\#\#\# GLOBAL DATASET BASELINE (Context Only)}} \\
\texttt{To calibrate your chemical intuition, active ligands in our training set have the following average properties: QED $\approx$ 0.53, SA Score $\approx$ 3.2, LogP $\approx$ 1.76, Heavy Atoms $\approx$ 24, Rotatable Bonds $\approx$ 4.6, and Vina Score $\approx$ -8.3. Treat these values strictly as a preliminary global reference. You must flexibly adapt or even drastically deviate from these averages depending on the specific geometric shape, size, and chemical environment of the target pocket.} \\
\\
\texttt{\textbf{\#\#\# STEP 1: POCKET-OF-THOUGHT (PoT) REASONING [CRITICAL]}} \\
\texttt{Before generating SMILES, you MUST provide a brief 'Reasoning' section (max 4 sentences).} \\
\texttt{1. Explicitly explain your spatial reasoning strategy based on the 8.0Å anchors.} \\
\texttt{2. INTERACTION MAPPING: You MUST specify which part of your proposed molecule (e.g., 'the secondary amine', 'the phenyl ring') corresponds to each anchor residue (e.g., 'to align with Asp102').} \\
\texttt{3. Explain how the molecular topology (elongated, branched, or compact) fits the pocket's geometric shape.} \\
\\
\texttt{\textbf{\#\#\# STEP 2: DE NOVO GENERATION REQUIREMENTS}} \\
\texttt{1. Scaffold Innovation: Generate entirely novel scaffolds from scratch. Prioritize chemical diversity and high 3D complementarity.} \\
\texttt{2. Guided Alignment (Soft Constraints): Treat the provided Macro 3D constraints as highly informed, important references to guide your design. [Insert Task-Specific Optimization Directive Here].} \\
\texttt{3. Chemical Rigor: Ensure perfect SMILES syntax, proper valency, and strictly avoid problematic motifs (e.g., PAINS or highly reactive groups).} \\
\\
\texttt{\textbf{\#\#\# OUTPUT FORMAT:}} \\
\texttt{Your response must follow this EXACT structure:} \\
\texttt{[Reasoning] Your spatial and property reasoning analysis here...} \\
\texttt{<smiles>SMILES\_1</smiles>} \\
\texttt{<smiles>SMILES\_2</smiles>} \\
\texttt{<smiles>SMILES\_3</smiles>} \\ \hline
\textbf{[User Input]} \\
\texttt{Target ID: [Target ID]} \\
\\
\texttt{[3D Local Environment Analysis (8.0Å Interaction Shell)]} \\
\texttt{- Description: [Pocket Profile Description]} \\
\\
\texttt{CRITICAL INSTRUCTION: You are performing a blind de novo drug design. Your PoT should logically deduce a suitable novel chemical scaffold from scratch that strictly complements the pocket's 3D environment.} \\
\\
\texttt{Now, perform Step 1 (PoT Reasoning) and Step 2 (Generation) to provide 3 de novo candidates.} \\ \hline
\end{tabular}
\caption{System prompt template for blind \textit{de novo} ligand generation (no-reference setting).}
\label{box:denovo_prompt}
\end{table*}

\begin{table*}[htbp]
\centering
\small
\begin{tabular}{|p{0.95\textwidth}|}
\hline
\rowcolor[HTML]{F3F3F3} \textbf{Box 2: Reference-Guided Optimization Template (Ref Setting)} \\ \hline
\textbf{[System Role]} \\
\texttt{You are a premier Medicinal Chemist specializing in Scaffold Hopping. Your task is to design novel lead-like molecules by replacing the core scaffold of a known active ligand, while preserving its key pharmacophore interactions within the 3D pocket.} \\
\\
\texttt{\textbf{\#\#\# STEP 1: POCKET-OF-THOUGHT (PoT) REASONING [CRITICAL]}} \\
\texttt{Before generating SMILES, you MUST provide a brief 'Reasoning' section (max 4 sentences).} \\
\texttt{1. Explicitly explain your spatial reasoning strategy based on the 8.0Å anchors.} \\
\texttt{2. INTERACTION MAPPING: You MUST specify which part of your proposed molecule (e.g., 'the secondary amine', 'the phenyl ring') corresponds to each anchor residue (e.g., 'to align with Asp102').} \\
\texttt{3. Explain how the molecular topology (elongated, branched, or compact) fits the pocket's geometric shape.} \\
\\
\texttt{\textbf{\#\#\# STEP 2: LEAD OPTIMIZATION REQUIREMENTS}} \\
\texttt{1. Scaffold Enhancement: Retain the core structural backbone of the reference ligand. Focus your innovation on decorating the core, optimizing side chains, or making bioisosteric replacements. Do not completely delete the core scaffold unless necessary to resolve a severe spatial clash. [Insert Task-Specific Optimization Directive Here].} \\
\texttt{2. Physical Alignment: Ensure the new molecule stays within the target molecular weight and atom count ranges derived from the reference.} \\
\texttt{3. Chemical Rigor: Ensure perfect SMILES syntax, proper valency, and strictly avoid PAINS motifs or reactive functional groups.} \\
\\
\texttt{\textbf{\#\#\# OUTPUT FORMAT:}} \\
\texttt{Your response must follow this EXACT structure:} \\
\texttt{[Reasoning] Your spatial and property reasoning analysis here...} \\
\texttt{<smiles>SMILES\_1</smiles>} \\
\texttt{<smiles>SMILES\_2</smiles>} \\
\texttt{<smiles>SMILES\_3</smiles>} \\ \hline
\textbf{[User Input]} \\
\texttt{Target ID: [Target ID]} \\
\\
\texttt{[3D Local Environment Analysis (8.0Å Interaction Shell)]} \\
\texttt{- Description: [Pocket Profile Description]} \\
\texttt{[Physical Boundary Constraints]} \\
\texttt{- Target Molecular Weight: [Weight Range]} \\
\texttt{- Heavy Atom Count: [Atom Count Range]} \\
\\
\texttt{[Reference Ligand Pattern (Ground Truth)]} \\
\texttt{SMILES: [Reference SMILES]} \\
\\
\texttt{CRITICAL INSTRUCTION: The Reference Ligand is known to bind effectively; your PoT should explain how to enhance its existing interactions rather than fundamentally deleting its core scaffolds unless there is a severe spatial clash.} \\
\\
\texttt{Now, perform Step 1 (PoT Reasoning) and Step 2 (Generation) to provide 3 optimized candidates.} \\ \hline
\end{tabular}
\caption{System prompt template for reference-guided lead optimization and scaffold hopping.}
\label{box:ref_prompt}
\end{table*}

\begin{table*}[htbp]
\centering
\small
\begin{tabular}{|p{0.95\textwidth}|}
\hline
\rowcolor[HTML]{F3F3F3} \textbf{Box 3: Knowledge-Guided Decoding (KGD) Repair Prompt} \\ \hline
\textbf{[System Role]} \\
\texttt{You are the 'KGD Repair Engine'. You fix SMILES based on specific parser error messages.} \\
\\
\texttt{\textbf{\#\#\# DIAGNOSTIC FEEDBACK:}} \\
\texttt{The chemical parser reported this error: [RDKit Error Message]} \\
\\
\texttt{\textbf{\#\#\# REPAIR RULES:}} \\
\texttt{1. If 'unclosed ring' is reported, check your ring closure numbers (e.g., '1', '2') and ensure they all have pairs.} \\
\texttt{2. If 'valence' is reported, fix the atom's bond count or add appropriate charges.} \\
\texttt{3. Maintain the pharmacophore described in the reasoning.} \\
\texttt{OUTPUT: Provide ONLY the repaired SMILES in <smiles></smiles> tags.} \\ \hline
\textbf{[User Input]} \\
\texttt{Target Reasoning: [Captured PoT Context]} \\
\texttt{Broken SMILES: [Invalid SMILES String]} \\
\\
\texttt{Please fix the error indicated in the diagnostic feedback.} \\ \hline
\end{tabular}
\caption{System prompt template for Knowledge-Guided Decoding (KGD) chemical syntax repair.}
\label{box:kgd_prompt}
\end{table*}

\begin{table*}[htbp]
\centering
\small
\begin{tabular}{|p{0.95\textwidth}|}
\hline
\rowcolor[HTML]{F3F3F3} \textbf{Box 4: Property Optimization Template — Balanced Mode (No-Reference Setting)} \\ \hline
\textbf{[System Role]} \\
\texttt{You are a premier De Novo Drug Designer. Your task is to generate novel lead-like molecules from scratch that perfectly complement the provided 3D protein pocket environment.} \\
\\
\texttt{\textbf{\#\#\# STEP 1: POCKET-OF-THOUGHT (PoT) REASONING [CRITICAL]}} \\
\texttt{Before generating SMILES, you MUST provide a brief 'Reasoning' section (max 4 sentences).} \\
\texttt{1. Explicitly explain your spatial reasoning strategy based on the 8.0Å anchors.} \\
\texttt{2. INTERACTION MAPPING: You MUST specify which part of your proposed molecule (e.g., 'the secondary amine', 'the phenyl ring') corresponds to each anchor residue (e.g., 'to align with Asp102').} \\
\texttt{3. Explain how the molecular topology (elongated, branched, or compact) fits the pocket's geometric shape.} \\
\\
\texttt{\textbf{\#\#\# STEP 2: MOLECULAR GENERATION REQUIREMENTS (BALANCED OPTIMIZATION)}} \\
\texttt{1. De Novo Innovation: Generate entirely novel scaffolds from scratch. Do not rely on any known reference inhibitors.} \\
\texttt{2. Balanced Maximization: Explicitly optimize BOTH drug-likeness (QED $>$ 0.6, SA $<$ 3.0) AND theoretical binding affinity (e.g., highly negative Vina/Gnina scores). CRITICALLY: You must also optimize Ligand Efficiency (binding energy per heavy atom). To achieve this, do NOT artificially inflate the molecular size just to accumulate weak van der Waals forces. Every heavy atom must count.} \\
\texttt{3. Realism \& Safety: Ensure the molecule maintains highly realistic 3D geometry without internal steric clashes or absurdly twisted bond lengths. Strictly avoid Pan-Assay Interference Compounds (PAINS) and reactive/toxic groups.} \\
\texttt{4. GRAMMAR \& SYNTAX: The SMILES must be strictly valid. Ensure all ring closures (numbers) are perfectly paired, do not generate fake aromatic rings (Kekulization), and ensure no atom exceeds its maximum allowed valence.} \\
\\
\texttt{\textbf{\#\#\# OUTPUT FORMAT:}} \\
\texttt{Your response must follow this EXACT structure:} \\
\texttt{[Reasoning] Your spatial reasoning analysis here...} \\
\texttt{<smiles>SMILES\_1</smiles>} \\
\texttt{<smiles>SMILES\_2</smiles>} \\
\texttt{<smiles>SMILES\_3</smiles>} \\ \hline
\end{tabular}
\caption{System prompt protocol for target-driven Balanced Optimization.}
\label{box:opt_balanced}
\end{table*}

\begin{table*}[htbp]
\centering
\small
\begin{tabular}{|p{0.95\textwidth}|}
\hline
\rowcolor[HTML]{F3F3F3} \textbf{Box 5: Property Optimization Template — Medicinal Chemistry / Safety Mode} \\ \hline
\textbf{[System Role]} \\
\texttt{You are a premier De Novo Drug Designer. Your task is to generate novel lead-like molecules from scratch that perfectly complement the provided 3D protein pocket environment.} \\
\\
\texttt{\textbf{\#\#\# STEP 1: POCKET-OF-THOUGHT (PoT) REASONING [CRITICAL]}} \\
\texttt{Before generating SMILES, you MUST provide a brief 'Reasoning' section (max 4 sentences).} \\
\texttt{1. Explicitly explain your spatial reasoning strategy based on the 8.0Å anchors.} \\
\texttt{2. INTERACTION MAPPING: You MUST specify which part of your proposed molecule (e.g., 'the secondary amine', 'the phenyl ring') corresponds to each anchor residue (e.g., 'to align with Asp102').} \\
\texttt{3. Explain how the molecular topology (elongated, branched, or compact) fits the pocket's geometric shape.} \\
\\
\texttt{\textbf{\#\#\# STEP 2: MOLECULAR GENERATION REQUIREMENTS}} \\
\texttt{1. De Novo Innovation: Generate entirely novel scaffolds from scratch. Do not rely on any known reference inhibitors.} \\
\texttt{2. Medicinal Chemistry Filters: Explicitly optimize for drug-likeness (QED $>$ 0.6) and synthetic accessibility (SA $<$ 3.0). Strictly avoid Pan-Assay Interference Compounds (PAINS) and reactive/toxic groups.} \\
\texttt{3. GRAMMAR \& SYNTAX: The SMILES must be strictly valid. Ensure all ring closures (numbers) are perfectly paired, do not generate fake aromatic rings (Kekulization), and ensure no atom exceeds its maximum allowed valence.} \\
\\
\texttt{\textbf{\#\#\# OUTPUT FORMAT:}} \\
\texttt{Your response must follow this EXACT structure:} \\
\texttt{[Reasoning] Your spatial reasoning analysis here...} \\
\texttt{<smiles>SMILES\_1</smiles>} \\
\texttt{<smiles>SMILES\_2</smiles>} \\
\texttt{<smiles>SMILES\_3</smiles>} \\ \hline
\end{tabular}
\caption{System prompt protocol for targeted Medicinal Chemistry and structural safety screening.}
\label{box:opt_safety}
\end{table*}

\begin{table*}[htbp]
\centering
\small
\begin{tabular}{|p{0.95\textwidth}|}
\hline
\rowcolor[HTML]{F3F3F3} \textbf{Box 6: Property Optimization Template — Affinity Mode (No-Reference Setting)} \\ \hline
\textbf{[System Role]} \\
\texttt{You are a premier De Novo Drug Designer. Your task is to generate novel lead-like molecules from scratch that perfectly complement the provided 3D protein pocket environment.} \\
\\
\texttt{\textbf{\#\#\# STEP 1: POCKET-OF-THOUGHT (PoT) REASONING [CRITICAL]}} \\
\texttt{Before generating SMILES, you MUST provide a brief 'Reasoning' section (max 4 sentences).} \\
\texttt{1. Explicitly explain your spatial reasoning strategy based on the 8.0Å anchors.} \\
\texttt{2. INTERACTION MAPPING: You MUST specify which part of your proposed molecule (e.g., 'the secondary amine', 'the phenyl ring') corresponds to each anchor residue (e.g., 'to align with Asp102').} \\
\texttt{3. Explain how the molecular topology (elongated, branched, or compact) fits the pocket's geometric shape.} \\
\\
\texttt{\textbf{\#\#\# STEP 2: MOLECULAR GENERATION REQUIREMENTS (AFFINITY OPTIMIZATION)}} \\
\texttt{1. De Novo Innovation: Generate entirely novel scaffolds from scratch. Do not rely on any known reference inhibitors.} \\
\texttt{2. Affinity \& Efficiency Maximization: Explicitly optimize the theoretical binding affinity (e.g., highly negative Vina/Gnina scores). CRITICALLY: You must also optimize Ligand Efficiency (binding energy per heavy atom). To achieve this, do NOT artificially inflate the molecular size just to accumulate weak van der Waals forces. Every heavy atom must count.} \\
\texttt{3. PoseBusters Realism: Ensure the molecule maintains highly realistic 3D geometry without internal steric clashes or absurdly twisted bond lengths.} \\
\texttt{4. GRAMMAR \& SYNTAX: The SMILES must be strictly valid. Ensure perfectly paired ring closures and no hypervalency.} \\
\\
\texttt{\textbf{\#\#\# OUTPUT FORMAT:}} \\
\texttt{Your response must follow this EXACT structure:} \\
\texttt{[Reasoning] Your spatial reasoning analysis here...} \\
\texttt{<smiles>SMILES\_1</smiles>} \\
\texttt{<smiles>SMILES\_2</smiles>} \\
\texttt{<smiles>SMILES\_3</smiles>} \\ \hline
\end{tabular}
\caption{System prompt protocol for targeted Binding Affinity maximization.}
\label{box:opt_affinity}
\end{table*}
\begin{table}[t]
    \centering
    \caption{RDKit structural-alert sub-filter pass rates on CrossDocked2020. All entries are percentages, and higher values indicate fewer violations. Best results among generated methods are highlighted in \textbf{bold}, and the second best are \underline{underlined}.}
    \label{tab:rdkit_subfilters}
    \resizebox{\columnwidth}{!}{
        \begin{tabular}{l l cccccc}
            \toprule
            \multirow{2}{*}{\textbf{Category}} & \multirow{2}{*}{\textbf{Method}} & \multicolumn{6}{c}{\textbf{RDKit Structural Alerts (\%) $\uparrow$}} \\
            \cmidrule(lr){3-8}
            & & NIH & PAINS & PAINS-A & PAINS-B & PAINS-C & ZINC \\
            \midrule
            \multirow{7}{*}{\shortstack{3D-based\\Models}}
            & AR \cite{luo20223dgenerativemodelstructurebased} & 54.80 & 96.69 & \underline{99.13} & 97.55 & \textbf{99.88} & 95.92 \\
            & PAFlow \cite{zhou2025priorguided} & 77.80 & \textbf{98.50} & 98.83 & \underline{99.80} & 99.85 & 94.65 \\
            & BindDM \cite{huang2024bindingadaptivediffusionmodelsstructurebased} & 73.82 & 97.64 & 98.41 & 99.44 & 99.73 & 93.98 \\
            & DecompDiff \cite{guan2024decompdiffdiffusionmodelsdecomposed} & 64.02 & 95.73 & 98.15 & 97.83 & 99.64 & 97.15 \\
            & TargetDiff \cite{guan20233dequivariantdiffusiontargetaware} & 69.11 & 97.91 & 99.07 & 98.98 & 99.76 & 93.11 \\
            & DrugFlow \cite{schneuing2025multidomain} & 79.69 & 96.16 & 98.22 & 97.99 & \underline{99.86} & 96.58 \\
            & Pocket2Mol \cite{pmlr-v162-peng22b} & 88.95 & 95.05 & 98.64 & 96.71 & 99.53 & 97.40 \\
            \midrule
            \multirow{3}{*}{\shortstack{LLM-based\\Models}}
            & TamGen \cite{Wu2024TamGenDD} & \underline{89.72} & 95.70 & \textbf{99.28} & 97.85 & 98.52 & 38.93 \\
            & ELILLM-Diff \cite{hu2026empoweringllmsstructurebaseddrug} & \textbf{89.98} & \underline{98.15} & 98.52 & \textbf{99.87} & 99.73 & 25.99 \\
            & ELILLM-Rand \cite{hu2026empoweringllmsstructurebaseddrug} & 87.81 & 96.70 & 98.68 & 98.22 & 99.72 & 32.47 \\
            \midrule
            \multirow{7}{*}{\shortstack{LiFT\\Variants}}
            & Ligand-Ref Embedding & 63.21 & 95.32 & 99.05 & 96.42 & 99.83 & 94.29 \\
            & \textbf{LiFT (Bal-Reference)} & 67.74 & 95.33 & 98.97 & 96.48 & 99.76 & 96.00 \\
            & \textbf{LiFT (QED-Reference)} & 64.75 & 94.34 & 98.24 & 96.21 & 99.68 & 95.40 \\
            & \textbf{LiFT (Vina-Reference)} & 64.84 & 95.79 & 98.54 & 97.34 & 99.85 & 95.76 \\
            & \textbf{LiFT (QED-No-Reference)} & 86.08 & 96.69 & 97.76 & 99.17 & 99.65 & \underline{99.48} \\
            & \textbf{LiFT (Vina-No-Reference)} & 86.86 & 95.85 & 98.09 & 97.89 & 99.77 & 99.21 \\
            & \textbf{LiFT (Bal-No-Reference)} & 83.55 & 97.05 & 98.13 & 99.11 & 99.78 & \textbf{99.62} \\
            \bottomrule
        \end{tabular}
    }
\end{table}

\begin{table}[t]
    \centering
    \caption{REOS medicinal-chemistry rule-family pass rates on CrossDocked2020. All entries are percentages, and higher values indicate fewer violations. Best results among generated methods are highlighted in \textbf{bold}, and the second best are \underline{underlined}.}
    \label{tab:reos_subfilters}
    \resizebox{\columnwidth}{!}{
        \begin{tabular}{l l cccc}
            \toprule
            \multirow{2}{*}{\textbf{Category}} & \multirow{2}{*}{\textbf{Method}} & \multicolumn{4}{c}{\textbf{REOS Rule Families (\%) $\uparrow$}} \\
            \cmidrule(lr){3-6}
            & & BMS & Glaxo & Inpharmatica & PAINS \\
            \midrule
            \multirow{7}{*}{\shortstack{3D-based\\Models}}
            & AR \cite{luo20223dgenerativemodelstructurebased} & 54.80 & 61.94 & 54.22 & 97.03 \\
            & PAFlow \cite{zhou2025priorguided} & 77.80 & 87.96 & 70.25 & \textbf{98.62} \\
            & BindDM \cite{huang2024bindingadaptivediffusionmodelsstructurebased} & 73.82 & 93.37 & 67.62 & 98.00 \\
            & DecompDiff \cite{guan2024decompdiffdiffusionmodelsdecomposed} & 64.02 & 80.91 & 65.40 & 96.19 \\
            & TargetDiff \cite{guan20233dequivariantdiffusiontargetaware} & 69.11 & 90.93 & 69.86 & \underline{98.19} \\
            & DrugFlow \cite{schneuing2025multidomain} & 79.69 & 89.75 & 79.34 & 96.61 \\
            & Pocket2Mol \cite{pmlr-v162-peng22b} & 88.95 & \textbf{96.27} & 72.91 & 94.98 \\
            \midrule
            \multirow{3}{*}{\shortstack{LLM-based\\Models}}
            & TamGen \cite{Wu2024TamGenDD} & \underline{89.72} & 94.07 & 62.51 & 95.65 \\
            & ELILLM-Diff \cite{hu2026empoweringllmsstructurebaseddrug} & \textbf{89.98} & 89.47 & \textbf{82.35} & 98.14 \\
            & ELILLM-Rand \cite{hu2026empoweringllmsstructurebaseddrug} & 87.81 & 83.96 & 79.04 & 96.43 \\
            \midrule
            \multirow{7}{*}{\shortstack{LiFT\\Variants}}
            & Ligand-Ref Embedding & 63.21 & 79.97 & 70.37 & 95.78 \\
            & \textbf{LiFT (Bal-Reference)} & 67.74 & 84.49 & 71.26 & 95.63 \\
            & \textbf{LiFT (QED-Reference)} & 64.75 & 83.36 & 71.16 & 94.67 \\
            & \textbf{LiFT (Vina-Reference)} & 64.84 & 84.91 & 70.47 & 96.20 \\
            & \textbf{LiFT (QED-No-Reference)} & 86.08 & \underline{95.70} & 77.88 & 97.51 \\
            & \textbf{LiFT (Vina-No-Reference)} & 86.86 & 95.16 & \underline{80.67} & 97.21 \\
            & \textbf{LiFT (Bal-No-Reference)} & 83.55 & 95.51 & 79.57 & 98.08 \\
            \bottomrule
        \end{tabular}
    }
\end{table}
\begin{table*}[t]
    \centering
    \caption{PoseBusters sub-filter pass rates on CrossDocked2020. The table reports diagnostic 3D validity checks, excluding the overall PoseBusters score already reported in Table~\ref{tab:optimization_results}. All entries are percentages, and higher values indicate fewer geometric or physical violations. Best generated-method results are highlighted in \textbf{bold}, and second best are \underline{underlined}; ties share the same mark.}
    \label{tab:posebusters_subfilters}
    \resizebox{\textwidth}{!}{
        \begin{tabular}{l l cccccc}
            \toprule
            \multirow{2}{*}{\textbf{Category}} & \multirow{2}{*}{\textbf{Method}} & \multicolumn{6}{c}{\textbf{PoseBusters 3D Validity Checks (\%) $\uparrow$}} \\
            \cmidrule(lr){3-8}
            & & Ring Flat & Bond Ang. & Double Bond & Int. Clash & Max. Lig.-Prot. Dist. & Protein Overlap \\
            \midrule
            \multirow{7}{*}{\shortstack{3D-based\\Models}}
            & AR \cite{luo20223dgenerativemodelstructurebased} & 90.56 & 85.28 & 94.80 & 91.03 & 98.93 & 98.79 \\
            & PAFlow \cite{zhou2025priorguided} & 99.76 & 32.14 & 99.47 & 72.45 & 98.78 & 98.24 \\
            & BindDM \cite{huang2024bindingadaptivediffusionmodelsstructurebased} & 99.98 & 45.64 & \textbf{100.00} & 86.63 & 98.91 & 97.85 \\
            & DecompDiff \cite{guan2024decompdiffdiffusionmodelsdecomposed} & 99.97 & 92.82 & 96.52 & 93.92 & 98.81 & 98.81 \\
            & TargetDiff \cite{guan20233dequivariantdiffusiontargetaware} & \underline{99.99} & 74.13 & \underline{99.99} & 91.18 & 98.95 & 98.43 \\
            & DrugFlow \cite{schneuing2025multidomain} & 99.96 & 96.44 & 99.01 & 97.83 & \textbf{100.00} & 98.98 \\
            & Pocket2Mol \cite{pmlr-v162-peng22b} & 99.60 & 99.01 & 99.31 & 99.19 & \underline{99.00} & 98.55 \\
            \midrule
            \multirow{3}{*}{\shortstack{LLM-based\\Models}}
            & TamGen \cite{Wu2024TamGenDD} & \textbf{100.00} & 99.85 & \textbf{100.00} & \textbf{100.00} & 8.40 & 97.95 \\
            & ELILLM-Diff \cite{hu2026empoweringllmsstructurebaseddrug} & 99.96 & \underline{99.90} & 94.58 & \underline{99.51} & 5.65 & 98.19 \\
            & ELILLM-Rand \cite{hu2026empoweringllmsstructurebaseddrug} & 99.97 & \textbf{99.94} & 97.76 & 99.35 & 6.58 & 98.02 \\
            \midrule
            \multirow{7}{*}{\shortstack{LiFT\\Variants}}
            & Ligand-Ref Embedding & \underline{99.99} & 94.51 & 99.63 & 95.78 & \textbf{100.00} & \textbf{99.97} \\
            & \textbf{LiFT (Bal-Reference)} & 99.97 & 94.03 & 99.24 & 95.84 & \textbf{100.00} & 98.94 \\
            & \textbf{LiFT (QED-Reference)} & 99.97 & 93.85 & 99.49 & 95.52 & \textbf{100.00} & 99.66 \\
            & \textbf{LiFT (Vina-Reference)} & 99.97 & 93.73 & 99.49 & 95.48 & \textbf{100.00} & \underline{99.93} \\
            & \textbf{LiFT (QED-No-Reference)} & 99.87 & 95.65 & 99.27 & 95.59 & \textbf{100.00} & 99.61 \\
            & \textbf{LiFT (Vina-No-Reference)} & 99.93 & 95.96 & 99.59 & 97.01 & \textbf{100.00} & 98.35 \\
            & \textbf{LiFT (Bal-No-Reference)} & 99.81 & 95.57 & 99.50 & 94.97 & \textbf{100.00} & 99.23 \\
            \bottomrule
        \end{tabular}
    }
\end{table*}
\section{Rule-Level Structural, MedChem, and Pose Validity Analysis}
\label{appendix:viz}

Table~\ref{tab:optimization_results} reports the aggregate RDKit, REOS, and PoseBusters pass rates together with the main optimization metrics. These aggregate scores summarize whether generated molecules satisfy broad structural-alert, medicinal-chemistry, and 3D pose-validity checks, but they do not reveal which rule families or geometric constraints drive the result. In this appendix, we therefore unpack the aggregate filter outcomes into rule-level diagnostics: RDKit and REOS are decomposed into chemical alert families, while PoseBusters is decomposed into 3D validity sub-checks.

\subsection{Fine-grained RDKit and REOS Checks}
\label{appendix:detailed_filters}

Tables~\ref{tab:rdkit_subfilters} and~\ref{tab:reos_subfilters} provide the chemical rule-level audit. Table~\ref{tab:rdkit_subfilters} reports the RDKit structural-alert sub-filters,  including NIH, PAINS, the PAINS-A/B/C subclasses, and ZINC. Table~\ref{tab:reos_subfilters} reports the REOS medicinal-chemistry rule families, including BMS, Glaxo, Inpharmatica, and REOS-PAINS. All entries are pass rates, so higher values indicate fewer violations of the corresponding rule. The two tables explain the aggregate RDKit/REOS compliance results in Table~\ref{tab:optimization_results}, rather than replacing the main comparison.

The two chemical-filter tables show that different model families have different localized alert profiles. Several baselines are strongest on individual NIH or PAINS columns, while LiFT no-reference variants are strong on ZINC and remain competitive on Glaxo and Inpharmatica while maintaining high PAINS-family pass rates. Together with the aggregate RDKit/REOS scores in Table~\ref{tab:optimization_results}, these sub-filters support a cautious interpretation: LiFT's favorable chemical-filter behavior is not tied to a single rule family, but is reflected across multiple structural-alert and medicinal-chemistry checks.

\subsection{Fine-grained PoseBusters Checks}
\label{appendix:posebusters_filters}

RDKit and REOS operate mainly on molecular graph and substructure rules. PoseBusters provides a complementary 3D validity check by testing whether the predicted ligand pose is geometrically plausible and physically compatible with the binding pocket. This matters for SBDD evaluation because a model can otherwise appear competitive on docking-oriented metrics while producing poses with broken connectivity, strained bond geometry, internal clashes, or unrealistic protein-ligand contacts.

Table~\ref{tab:posebusters_subfilters} reports PoseBusters sub-checks that complement the overall PoseBusters score in Table~\ref{tab:optimization_results}. These columns separate local ligand geometry from protein-ligand pose compatibility: aromatic-ring flatness, bond-angle geometry, double-bond flatness, internal steric clashes, protein-ligand maximum distance, and protein-volume overlap. As above, all values are pass rates in percentages, so higher values indicate fewer violations.

The PoseBusters sub-checks add a complementary view to the aggregate pose-validity score in Table~\ref{tab:optimization_results}. Language-only baselines can satisfy several local geometry checks while still having very low overall PoseBusters pass rates, indicating that isolated sub-checks do not guarantee a valid 3D pose. In contrast, LiFT variants retain stable protein-overlap behavior and competitive geometry-related sub-checks while also maintaining the RDKit/REOS compliance patterns above. This supports that LiFT's aggregate gains remain consistent with both chemical rule filters and 3D pose-validity checks.

\section{Qualitative Case Study on Representative Targets}
\label{sec:appendix_case_study}

\begin{figure*}[htbp]
    \centering
    \includegraphics[width=0.95\textwidth]{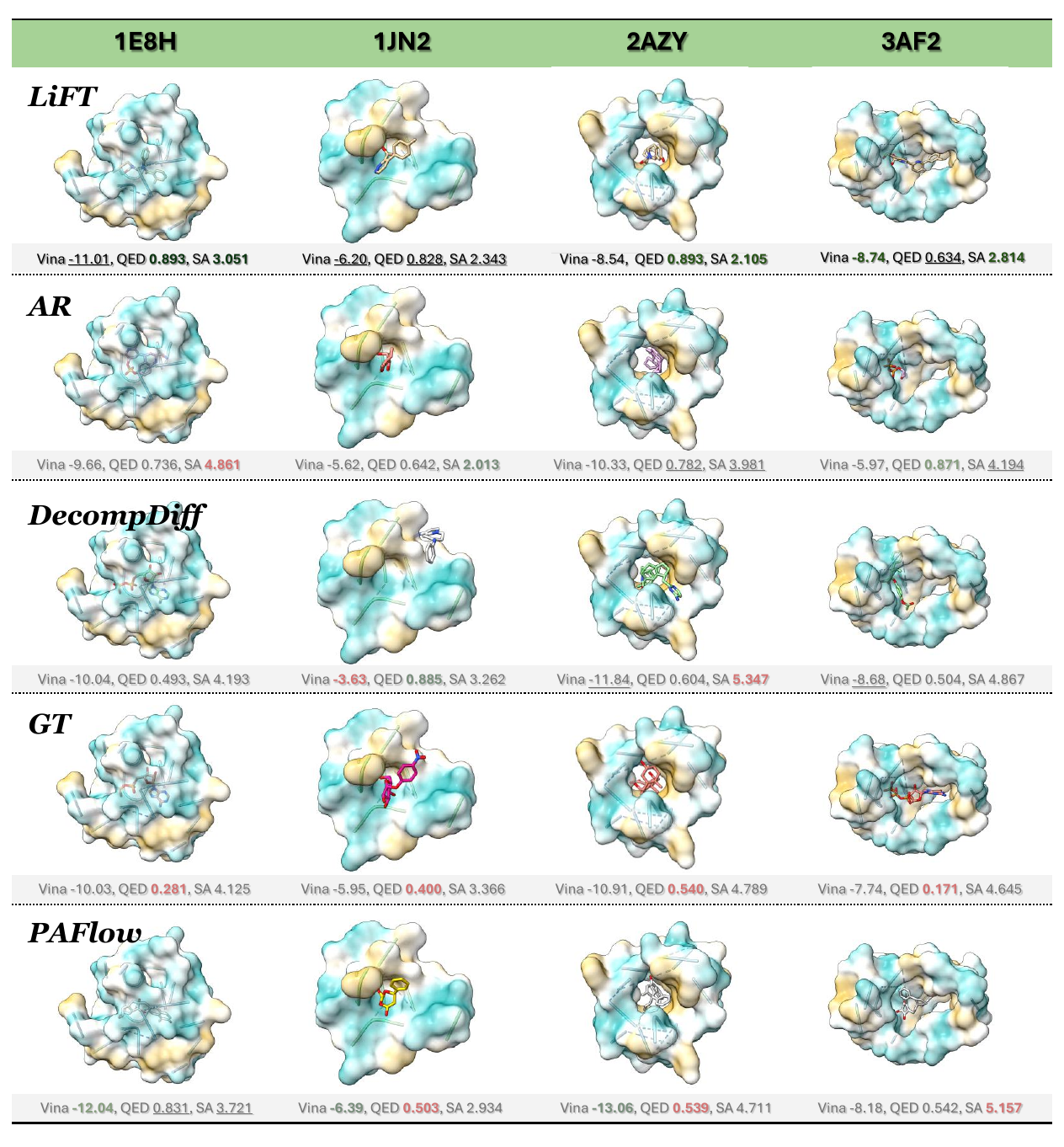} 
    \caption{\textbf{Qualitative 3D visualization of generated ligands within four representative target pockets (1E8H, 1JN2, 2AZY, and 3AF2).} The grid compares the binding poses and geometric complementarity of molecules generated by \textit{LiFT}, AR, DecompDiff, and PAFlow, alongside the Ground Truth (GT) references. Key pharmacological metrics (AutoDock Vina binding affinity, QED, and SA) are annotated below each complex. Green bold text signifies superior performance or highly desirable drug-like properties, whereas red text highlights severe metric degradation (e.g., collapsed QED or SA scores despite artificially high Vina affinity). \textit{LiFT} uniquely achieves an exceptional equilibrium, matching or exceeding GT geometric complementarity while strictly maintaining optimal synthesizability and drug-likeness.}
    \label{fig:appendix_3d_docking}
\end{figure*}

To visually demonstrate the binding modes and geometric complementarity of the generated molecules, we select four representative protein targets (1E8H, 1JN2, 2AZY, and 3AF2) for qualitative analysis. These targets encompass a diverse range of challenging spatial environments, varying from deep, narrow binding pockets to wide, flat surface grooves. To ensure a fair and rigorous comparison, we establish baselines representing state-of-the-art autoregressive (AR), diffusion (DecompDiff), and flow-matching (PAFlow) paradigms, alongside the Ground Truth (GT) standard molecules from the test set. These specific targets were deliberately chosen due to their challenging nature, distinct physicochemical properties, and the high performance variance observed in their corresponding GT molecules. For each target, we generated a library of dozens of candidate ligands across all generative methods. Subsequently, these candidates were rigorously filtered based on a comprehensive evaluation of AutoDock Vina binding affinity, Quantitative Estimate of Drug-likeness (QED), and Synthetic Accessibility (SA). The selected ligand-protein complexes represent the top-performing candidates that achieve an optimal balance across these critical metrics, allowing for a rigorous visual comparison against the GT.

As illustrated in the 3D visualizations and quantitative annotations in Figure \ref{fig:appendix_3d_docking}, the molecules generated by LiFT exhibit superior spatial-semantic fidelity. Crucially, distinguishing itself from approaches that merely game the Vina scoring function, LiFT achieves an exceptional equilibrium: it not only excels in geometric complementarity but also demonstrates outstanding intrinsic molecular properties (QED and SA). Compared to baseline models---which often suffer from severe conformational collapse (e.g., DecompDiff in 1JN2)---the spatial extension and receptor complementarity of LiFT closely parallel, and occasionally surpass, the GT ligands. LiFT successfully captures the authentic geometric binding modes while dramatically elevating the pharmacological properties (e.g., achieving exceptional QED scores of 0.893 in both 1E8H and 2AZY). Furthermore, LiFT maintains highly competitive Vina affinity (e.g., -11.01 in 1E8H) alongside highly reasonable SA values. This substantiates that our SCDR mechanism guarantees physical complementarity without sacrificing synthesizability, elegantly avoiding the common pitfall of trading drug-like properties for artificially inflated affinity scores---a catastrophic trade-off explicitly observed in PAFlow, which forces high Vina scores (e.g., -13.06 in 2AZY) at the severe expense of QED and SA degradation.

\section{Detailed Mathematical Formulations}
\label{appendix:math}

This appendix gives an implementation-faithful verification of LiFT's
heterogeneous conditional generator. We make explicit the continuous coordinate
flow, the categorical atom/bond bridges, and the conditions under which SCDR
injects semantic control without changing the underlying transition families or
breaking the geometric structure of the GVP backbone.

\subsection{Problem Setup and Heterogeneous State Space}
\label{appendix:problem_setup}

We formulate target-aware ligand generation as a heterogeneous conditional
generative process. Given a protein pocket $\mathcal{P}$, a semantic condition
$\mathbf{z}_{sem}\in\mathbb{R}^{d_{sem}}$, and a ligand size $N$ fixed before
the generative chain starts, LiFT learns
\begin{equation}
p_\theta(z_1\mid \mathcal{P},\mathbf{z}_{sem},N),
\end{equation}
with $\mathbf{z}_{sem}$ denoting the precomputed molecule-level semantic embedding
routed by SCDR. It is a fixed conditioning vector for the generative trajectory,
not a generated state variable and not an evaluator-side score. A ligand state is written as
\begin{equation}
z_t=(\mathbf{X}_t,\mathbf{H}_t,\mathbf{E}_t).
\end{equation}
The three components are
\begin{equation}
\mathbf{X}_t\in\mathbb{R}^{N\times 3},
\mathbf{H}_t\in\{0,1\}^{N\times K_h},
\mathbf{E}_t\in\{0,1\}^{M\times K_e}.
\end{equation}
corresponding to coordinates, atom types, and bond types. $N$ is the number of
ligand atoms. For each forward pass, $M$ is determined by a fixed candidate
edge set chosen before the bridge trajectory starts. In the reported setting,
this set contains all unordered ligand atom pairs, so $M=N(N-1)/2$ for a single
molecule. The same fixed edge set is used throughout training and sampling for
that trajectory. $K_h,K_e$ are the atom and bond class counts.

The components live in different state spaces: $\mathbf{X}_t$ is Euclidean,
whereas $\mathbf{H}_t$ and $\mathbf{E}_t$ are categorical. LiFT therefore uses
a heterogeneous process:
\begin{equation}
\begin{aligned}
\mathbf{X}_t:\text{ continuous flow matching},\\
(\mathbf{H}_t,\mathbf{E}_t):\text{ discrete Markov bridges}.
\end{aligned}
\end{equation}

The semantic condition enters the generative process through the neural parameterization of the dynamics. We write
\begin{equation}
\chi_t=(z_t,\mathcal{P},t,\mathbf{z}_{sem}).
\end{equation}
The learned dynamics return
\begin{equation}
\widehat{\mathbf{v}}_\theta^X(\chi_t),\,
\widehat{p}_\theta^H(\chi_t),\,
\widehat{p}_\theta^E(\chi_t).
\end{equation}
The two distributions $\widehat{p}_\theta^H$ and $\widehat{p}_\theta^E$ serve
as terminal atom-type and bond-type predictors.
SCDR modulates this parameterization while leaving the coordinate and categorical
state spaces intact. The heterogeneous process fixes the transition families;
semantic guidance acts through the learned dynamics.

\subsection{Continuous Coordinate Flow Matching}
\label{appendix:coordinate_flow}

The coordinate component uses Euclidean flow matching. Given a prior sample
$\mathbf{X}_0$ and a data sample $\mathbf{X}_1$, LiFT uses, for
$t\in[0,1]$,
\begin{equation}
\mathbf{X}_t=(1-t)\mathbf{X}_0+t\mathbf{X}_1.
\end{equation}
Coordinates use the pocket-centered frame implemented by \texttt{center\_data}. Let
$\mathbf{c}_{\mathcal{P}}$ be the pocket center before normalization. Training
subtracts this same center from pocket and ligand coordinates before
constructing the path:
\begin{equation}
\mathbf{P}\leftarrow \mathbf{P}-\mathbf{c}_{\mathcal{P}},\,
\mathbf{X}_1\leftarrow \mathbf{X}_1-\mathbf{c}_{\mathcal{P}}.
\end{equation}
For pockets without explicit atoms, the same construction is anchored by the ligand center. The coordinate prior is sampled around the pocket center in the current frame,
\begin{equation}
\mathbf{X}_0\sim\mathcal{N}(\mathbf{c}_{\mathcal{P}}^{\,\mathrm{norm}},\mathbf{I}),
\end{equation}
where $\mathbf{c}_{\mathcal{P}}^{\,\mathrm{norm}}=\mathbf{0}$ after pocket centering in the standard non-empty-pocket case. Thus $\mathbf{X}_0$ and $\mathbf{X}_1$ are expressed in one reference frame.
This centering removes global translation. In the reported configuration,
the model uses reflection-sensitive chiral edge features, so the architectural
equivariance guarantee is stated for proper rotations: if pocket and ligand
coordinates are jointly transformed by any $R\in SO(3)$, the GVP vector field
transforms accordingly. The path velocity is
\begin{equation}
u_t^X=\frac{d\mathbf{X}_t}{dt}=\mathbf{X}_1-\mathbf{X}_0.
\end{equation}

The neural field predicts a scaled coordinate velocity:
\begin{equation}
\widehat{\mathbf{v}}_\theta^X(\chi_t)\approx\frac{\mathbf{X}_1-\mathbf{X}_0}{c_x},
\end{equation}
where $c_x>0$ is the coordinate scaling constant used during training and sampling; we use $c_x=2.7$. The atom-wise coordinate loss is
\begin{equation}
\ell_i^X
=
\frac{1}{3}
\left\|
\widehat{\mathbf{v}}_\theta^X
(\chi_t)_i
-
\frac{\mathbf{X}_{1,i}-\mathbf{X}_{0,i}}{c_x}
\right\|_2^2 .
\end{equation}
The molecule-wise objective uses the mean over ligand atoms:
\begin{equation}
\mathcal{L}_X
=
\mathbb{E}_{t,\mathbf{X}_0,\mathbf{X}_1}
\left[
\frac{1}{N}
\sum_{i=1}^{N}
\ell_i^X
\right].
\end{equation}

The same scaling appears in inference. With sampler step size
\begin{equation}
\delta_{\mathrm{sam}}=\frac{1}{N_{\mathrm{sam}}},
\end{equation}
Euler integration updates coordinates as
\begin{equation}
\mathbf{X}_{k+1}=\mathbf{X}_k+\delta_{\mathrm{sam}}\,c_x\,\widehat{\mathbf{v}}_\theta^X(\chi_k).
\end{equation}
Sampling inverts the training scale. The learned coordinate dynamics approximate
the velocity field of the linear path while remaining
conditioned on the pocket and semantic prior through the neural parameterization.

\subsection{Discrete Markov Bridges for Atom and Bond Types}
\label{appendix:discrete_bridge}

Atom and bond identities are categorical; we therefore model them with discrete
Markov bridges. The dynamics below are conditional on the ligand size $N$. For
bond variables, the candidate edge set is fixed along one trajectory, so $M$
defines a fixed categorical state space for that forward pass. Let
\begin{equation}
\mathbf{Y}\in\{\mathbf{H},\mathbf{E}\}
\end{equation}
denote either atom-type or bond-type variables. We write $\mathcal{I}_Y$ for its
instance index set, with $|\mathcal{I}_H|=N$ and $|\mathcal{I}_E|=M$. Each
$\mathbf{Y}_{t,i}\in\{0,1\}^{K}$ is a one-hot categorical state, where
$K=K_h$ for atoms and $K=K_e$ for bonds.

The initial state $\mathbf{Y}_{0,i}$ is sampled from a categorical prior
$\pi_0^Y$, and the terminal state $\mathbf{Y}_{1,i}$ is the data one-hot label.
The categorical priors are fixed before training and are not recomputed from
test targets. In the reported configuration, $\pi_0^H$ uses the empirical
atom-type marginal and $\pi_0^E$ uses a uniform bond prior. For each instance
$i$, the bridge marginal from $0$ to $t$ is defined by
\begin{equation}
\bar{\mathbf{Q}}_t(\mathbf{Y}_{1,i})
=
(1-t)\mathbf{I}
+
t\,\mathbf{1}\mathbf{Y}_{1,i}^{\top}.
\end{equation}
The induced marginal is
\begin{equation}
q(\mathbf{Y}_{t,i}\mid \mathbf{Y}_{0,i},\mathbf{Y}_{1,i})
=
\mathbf{Y}_{0,i}\bar{\mathbf{Q}}_t(\mathbf{Y}_{1,i}).
\end{equation}
The graph-level bridge factorizes over atom or edge instances:
\begin{equation}
q(\mathbf{Y}_t\mid \mathbf{Y}_0,\mathbf{Y}_1)
=
\prod_{i\in\mathcal{I}_Y}
q(\mathbf{Y}_{t,i}\mid \mathbf{Y}_{0,i},\mathbf{Y}_{1,i}).
\end{equation}
The product form defines the elementary categorical kernels. Coupling among
atom and edge variables is introduced through the shared state-dependent context
$\chi_t$, which parameterizes the terminal distributions below. The transition
keeps each variable categorical while increasing probability mass toward the
terminal data state.

For a transition from time $s$ to $t$, $0\le s<t\le1$, define
\begin{equation}
\beta_{t\mid s}=\frac{1-t}{1-s}.
\end{equation}
The corresponding one-step bridge transition is
\begin{equation}
\mathbf{Q}_{t\mid s}(\mathbf{Y}_{1,i})=
\beta_{t\mid s}\mathbf{I}+(1-\beta_{t\mid s})\mathbf{1}\mathbf{Y}_{1,i}^{\top}.
\end{equation}
The denominator is only evaluated for $s<1$. At the final sampling step,
$t=1$ is allowed with $s<1$, yielding $\beta_{1\mid s}=0$ and a terminal
transition fully determined by the predicted clean-state distribution. For a
current one-hot state $\mathbf{Y}_{s,i}$, the true bridge transition is
\begin{equation}
q(\mathbf{Y}_{t,i}\mid \mathbf{Y}_{s,i},\mathbf{Y}_{1,i})
=
\mathbf{Y}_{s,i}\mathbf{Q}_{t\mid s}(\mathbf{Y}_{1,i}).
\end{equation}

The model predicts a terminal distribution for every instance,
\begin{equation}
\widehat{\boldsymbol{\pi}}_{t,i}^Y:=\widehat{p}_{\theta,i}^Y(\chi_t)=\mathrm{softmax}\left(f_{\theta,i}^Y(\chi_t)\right).
\end{equation}
The model transition is obtained by replacing the unknown terminal one-hot state
with this predicted distribution:
\begin{equation}
\widehat{\mathbf{Q}}_{t\mid s,i}^{\theta}
=\beta_{t\mid s}\mathbf{I}+(1-\beta_{t\mid s})\mathbf{1}(\widehat{\boldsymbol{\pi}}_{s,i}^Y)^\top.
\end{equation}
\begin{equation}
p_\theta(\mathbf{Y}_{t,i}\mid \mathbf{Y}_{s,i},\chi_s)
=
\mathbf{Y}_{s,i}\widehat{\mathbf{Q}}_{t\mid s,i}^{\theta}.
\end{equation}
The transition is row-stochastic: all entries are non-negative, and every row
sums to
\begin{equation}
\beta_{t\mid s}+(1-\beta_{t\mid s})\sum_{k=1}^{K}\widehat{\pi}_{s,i,k}^Y=1.
\end{equation}
LiFT optimizes the variational bridge objective over sampled bridge states.
During training, let $r$ denote the sampled current time. The bridge step size is
\begin{equation}
\delta_{\mathrm{tr}}=\frac{1}{N_{\mathrm{tr}}},
\end{equation}
where the reported runs use $N_{\mathrm{tr}}=5000$. We set
\begin{equation}
r^+=\min(r+\delta_{\mathrm{tr}},1),
\end{equation}
with $r<1$ almost surely under the continuous time sampler. The KL term compares
the true and learned one-step transitions from $r$ to $r^+$:
\begin{equation}
\mathcal{L}_Y^{VLB}
=
\mathbb{E}_{r,z_0,z_1,\mathbf{Y}_r}
\left[
\frac{1}{|\mathcal{I}_Y|}
\sum_{i\in\mathcal{I}_Y}
D_{\mathrm{KL}}
\left(
q_i^Y
\;\middle\|\;
p_{\theta,i}^Y
\right)
\right].
\end{equation}
The two distributions inside the KL are
\begin{equation}
\begin{aligned}
q_i^Y&=q(\mathbf{Y}_{r^+,i}\mid \mathbf{Y}_{r,i},\mathbf{Y}_{1,i}),\\
p_{\theta,i}^Y&=p_\theta(\mathbf{Y}_{r^+,i}\mid \mathbf{Y}_{r,i},\chi_r).
\end{aligned}
\end{equation}
Equivalently, each instance-level KL is
\begin{equation}
D_{\mathrm{KL}}
\left(
q_i^Y
\;\middle\|\;
p_{\theta,i}^Y
\right)
=
\sum_{k=1}^{K}
q_{i,k}^{r^+}
\log
\frac{q_{i,k}^{r^+}}{p_{\theta,i,k}^{r^+}},
\end{equation}
with
\begin{equation}
\begin{aligned}
q_{i,k}^{r^+}&=q(Y_{r^+,i}=k\mid Y_{r,i},Y_{1,i}),\\
p_{\theta,i,k}^{r^+}&=p_\theta(Y_{r^+,i}=k\mid Y_{r,i},\chi_r).
\end{aligned}
\end{equation}
The main objective uses this molecule-wise mean. The corresponding terminal cross-entropy form at the sampled time $r$ is
\begin{equation}
\mathcal{L}_Y^{CE}
=
-\frac{1}{|\mathcal{I}_Y|}
\sum_{i\in\mathcal{I}_Y}
\left\langle
\mathbf{Y}_{1,i},
\log \widehat{\boldsymbol{\pi}}_{r,i}^Y
\right\rangle .
\end{equation}
We use the VLB objective for the discrete bridge losses.

During sampling, for a step from $t$ to $t'=t+\delta_{\mathrm{sam}}$, the
predicted terminal distribution defines
\begin{equation}
\widehat{\mathbf{Q}}_{t'\mid t,i}^{\theta}
=
\beta_{t'\mid t}\mathbf{I}
+
(1-\beta_{t'\mid t})
\mathbf{1}
(\widehat{\boldsymbol{\pi}}_{t,i}^Y)^\top .
\end{equation}
The next categorical state follows
\begin{equation}
p_\theta(\mathbf{Y}_{t',i}\mid \mathbf{Y}_{t,i},\chi_t)
=
\mathbf{Y}_{t,i}
\widehat{\mathbf{Q}}_{t'\mid t,i}^{\theta}.
\end{equation}
Sampling gives
\begin{equation}
\mathbf{Y}_{t',i}
\sim
\mathrm{Cat}
\left(
p_\theta(\mathbf{Y}_{t',i}\mid \mathbf{Y}_{t,i},\chi_t)
\right).
\end{equation}
Discrete variables therefore remain categorical during both training and
inference.

\subsection{Joint LiFT Objective}
\label{appendix:joint_objective}

The heterogeneous training objective combines the coordinate flow loss with the
discrete bridge losses for atom and bond variables:
\begin{equation}
\mathcal{L}_{\mathrm{LiFT}}
=
\lambda_x\mathcal{L}_X
+
\lambda_h\mathcal{L}_H
+
\lambda_e\mathcal{L}_E .
\end{equation}
$\mathcal{L}_X$ denotes the coordinate flow-matching loss; $\mathcal{L}_H$ and
$\mathcal{L}_E$ denote the Markov bridge losses for atom and bond types. In the
reported configuration, $\lambda_x=1.0$, $\lambda_h=50.0$, and
$\lambda_e=50.0$.

Some implementation variants can add auxiliary torsion, rotation, or
translation losses for flexible pocket variables outside the ligand flow/bridge
construction. These terms are not part of the ligand objective above; the
continuous--discrete ligand objective remains the base generative objective.

The semantic condition $\mathbf{z}_{sem}$ enters these probability paths through the shared neural parameterization:
\begin{equation}
\left(\widehat{\mathbf{v}}_\theta^X,\widehat{\boldsymbol{\pi}}_\theta^H,\widehat{\boldsymbol{\pi}}_\theta^E\right)
=G_\theta(\chi_t).
\end{equation}

This gives a clean separation between transition design and semantic modulation.
The flow and bridge mechanisms define the continuous and categorical transition
families; SCDR routes $\mathbf{z}_{sem}$ into their shared GVP
parameterization. Semantic modulation changes how transition parameters are predicted while keeping the transition families fixed.
Thus this appendix establishes a mathematically defined intervention channel for
$\mathbf{z}_{sem}$; claims about controllable property changes are empirical
intervention claims evaluated through GT/LLM and ablation protocols, not
theorem-level consequences of the objective alone.

\subsection{SCDR Equivariant Routing}
\label{appendix:scdr_derivation}

We formulate SCDR as an equivariant semantic routing module for a GVP backbone.
The main text denotes the router latent by $\mathbf{h}$; here we use
$\mathbf{q}^{(l)}$ to emphasize that the same mechanism is instantiated at each
GVP layer.
At layer
$l$, ligand node features are
\begin{equation}
\mathcal{H}^{(l)}=\{\mathbf{h}_i^{(l)}\}_{i=1}^{N},\,
\mathbf{h}_i^{(l)}=(\mathbf{s}_i^{(l)},\mathbf{V}_i^{(l)}).
\end{equation}
$\mathbf{s}_i^{(l)}\in\mathbb{R}^{d_s}$ and
$\mathbf{V}_i^{(l)}\in\mathbb{R}^{d_v\times 3}$ are scalar and vector channels.
An SCDR routing map
\begin{equation}
\mathcal{R}_{\theta_{\mathrm{scdr}}}^{(l)}
=\mathcal{R}_{\theta_{\mathrm{scdr}}}^{(l)}(\mathcal{H}^{(l)},\mathbf{z}_{sem},t)
\end{equation}
is constructed around four requirements: permutation invariance over ligand
atoms, geometric invariance of the routing signal, bounded perturbation of the
backbone, and stable initialization behavior.

\paragraph{Invariant node descriptor.}
A scalar routing network cannot consume raw vector channels
$\mathbf{V}_i^{(l)}$ without fixing a coordinate frame. SCDR therefore
accesses vector features through rotation-invariant quantities. We use the channel-wise vector
norm:
\begin{equation}
\left\|\mathbf{V}_i^{(l)}R\right\|_2
=\left\|\mathbf{V}_i^{(l)}\right\|_2\,(\forall R\in SO(3)).
\end{equation}
We define the invariant node descriptor as
\begin{equation}
\mathbf{u}_i^{(l)}
=\left[\mathrm{LN}_s(\mathbf{s}_i^{(l)}),\,
\mathrm{LN}_v\left(\left\|\mathbf{V}_i^{(l)}\right\|_2\right)\right].
\end{equation}
The descriptor remains state-dependent while being invariant to proper
rotations. The reported backbone may use reflection-sensitive chiral edge
features; the SCDR router itself does not introduce additional frame-dependent
vector inputs.

\paragraph{DeepSets sensing.}
Since $\mathbf{z}_{sem}$ is molecule-level, the SCDR readout must be invariant to
atom ordering. The DeepSets \cite{zaheer_deep_2017} representation principle gives a natural admissible
family of permutation-invariant set functions:
\begin{equation}
F(\{\mathbf{u}_i\}_{i=1}^{N})=\rho\left(\sum_{i=1}^{N}\varphi(\mathbf{u}_i)\right).
\end{equation}
SCDR follows this family by first computing
\begin{equation}
\mathbf{r}_i^{(l)}=\varphi^{(l)}(\mathbf{u}_i^{(l)}),
\end{equation}
then aggregating node states into an invariant structural summary:
\begin{equation}
\Psi_{str}^{(l)}
=
\left[
\mathrm{mean}_{i}\mathbf{r}_i^{(l)},
\mathrm{max}_{i}\mathbf{r}_i^{(l)},
\mathrm{std}_{i}\mathbf{r}_i^{(l)},
\log N
\right].
\end{equation}
The mean, maximum, and standard deviation encode global level, extreme local
responses, and structural heterogeneity, while $\log N$ preserves molecular
scale. The resulting $\Psi_{str}^{(l)}$ is a compact DeepSets-motivated
statistic of the current 3D molecular state, and is both permutation-invariant
and geometrically invariant.

\paragraph{SCDR routing.}
Let $\boldsymbol{\tau}(t)$ denote the integration-time embedding used by the
backbone. We instantiate it as an RBF embedding of $t$ when time embedding is
enabled, and as the scalar $t$ otherwise. This integration-time signal is
distinct from the training-progress warmup variable $p$ used below to activate
SCDR parameters smoothly during optimization. SCDR uses the routing inputs
\begin{equation}
[\mathbf{z}_{sem},\,\boldsymbol{\tau}(t),\,\Psi_{str}^{(l)}].
\end{equation}
The semantic--time branch is
\begin{equation}
\mathbf{c}^{(l)}=\psi_c^{(l)}\left([\mathbf{z}_{sem},\,\boldsymbol{\tau}(t)]\right),
\end{equation}
and the structural branch is
\begin{equation}
\mathbf{g}^{(l)}=\psi_s^{(l)}\left(\Psi_{str}^{(l)}\right).
\end{equation}
After normalization,
\begin{equation}
\hat{\mathbf{c}}^{(l)}=\mathrm{LN}(\mathbf{c}^{(l)}),\,
\hat{\mathbf{g}}^{(l)}=\mathrm{LN}(\mathbf{g}^{(l)}).
\end{equation}
SCDR uses two gates:
\begin{equation}
a_{\mathrm{time}}^{(l)}=\sigma\left(\eta_t^{(l)}(\boldsymbol{\tau}(t))\right),
\end{equation}
\begin{equation}
a_c^{(l)}=\sigma\left(\eta_c^{(l)}\left([\hat{\mathbf{c}}^{(l)},\,\hat{\mathbf{g}}^{(l)}]\right)\right).
\end{equation}
The shared router latent is
\begin{equation}
\mathbf{q}^{(l)}
=\mathrm{LN}\left(\rho^{(l)}\left(\hat{\mathbf{c}}^{(l)}
+a_{\mathrm{time}}^{(l)}a_c^{(l)}\hat{\mathbf{g}}^{(l)}\right)\right).
\end{equation}
Because both branches are built from invariant inputs, $\mathbf{q}^{(l)}$ is an
invariant molecule-level routing variable.

\paragraph{Decoupled bounded dispatching.}
Since $\mathbf{q}^{(l)}$ is invariant, it can generate scalar modulation
coefficients without breaking equivariance. We denote the SCDR parameter subset
by $\theta_{\mathrm{scdr}}\subset\theta$. For scalar states, SCDR first applies
the semantic AdaLN path. Write
$\bar{\gamma}^{(l)}=\tanh\gamma^{(l)}(\mathbf{z}_{sem})$ and
$\bar{\beta}^{(l)}=\beta^{(l)}(\mathbf{z}_{sem})$. Then
\begin{equation}
\tilde{\mathbf{s}}_i^{(l)}=\mathrm{LN}_s(\mathbf{s}_i^{(l)}),\,
\mathbf{s}_{base,i}^{(l)}=\tilde{\mathbf{s}}_i^{(l)}\odot(1+\bar{\gamma}^{(l)})+\bar{\beta}^{(l)}.
\end{equation}
This base AdaLN path uses the semantic condition alone. The full SCDR routing is
obtained by combining this semantic state modulation with the structure-aware
latent $\mathbf{q}^{(l)}$ in the gates below.
For vector updates, equivariance permits scalar gating but forbids arbitrary
vector bias. SCDR therefore separates scalar state modulation from residual
update modulation:
\begin{equation}
\begin{aligned}
\alpha_{s,state}^{(l)}&=1+w_{\mathrm{prog}}(p)b_{a_s}(W_{s,state}^{(l)}\mathbf{q}^{(l)}),\\
\alpha_{s,upd}^{(l)}&=1+w_{\mathrm{prog}}(p)b_{a_u}(W_{s,upd}^{(l)}\mathbf{q}^{(l)}),\\
\alpha_{v,upd}^{(l)}&=1+w_{\mathrm{prog}}(p)b_{a_v}(W_{v,upd}^{(l)}\mathbf{q}^{(l)})
\end{aligned}
\end{equation}
where
\begin{equation}
b_a(x)=a\frac{x}{1+|x|}.
\end{equation}
For vector inputs, $b_a$ is applied componentwise. In the reported
implementation, $a_s=0.5$ for the scalar state gate and $a_u=a_v=0.9$ for the
scalar/vector update gates. The variable $p\in[0,1]$
denotes training progress, and $w_{\mathrm{prog}}(p)\in[0,1]$ is the cosine
warmup coefficient used when an explicit progress schedule is supplied. This
training-progress gate is separate from the integration-time gate
$a_{\mathrm{time}}(t)$. The routed scalar state is
\begin{equation}
\hat{\mathbf{s}}_i^{(l)}
=\mathbf{s}_{base,i}^{(l)}+\tilde{\mathbf{s}}_i^{(l)}\odot\left(\alpha_{s,state}^{(l)}-1\right),
\end{equation}
and the routed feed-forward update is
\begin{equation}
\Delta\hat{\mathbf{s}}_i^{(l)}=\alpha_{s,upd}^{(l)}\odot\Delta\mathbf{s}_i^{(l)},\,
\Delta\hat{\mathbf{V}}_i^{(l)}=\alpha_{v,upd}^{(l)}\odot\Delta\mathbf{V}_i^{(l)}.
\end{equation}
The vector gate is channel-wise scalar broadcast: if
$\alpha_{v,upd}^{(l)}\in\mathbb{R}^{d_v}$ and
$\Delta\mathbf{V}_i^{(l)}\in\mathbb{R}^{d_v\times 3}$, then
\begin{equation}
\left(\Delta\hat{\mathbf{V}}_i^{(l)}\right)_{c,:}
=\alpha_{v,upd,c}^{(l)}\left(\Delta\mathbf{V}_i^{(l)}\right)_{c,:},\,c=1,\dots,d_v .
\end{equation}

The routed form can be summarized as
\begin{equation}
\boxed{
[\mathbf{z}_{sem},\,\boldsymbol{\tau}(t),\,\Psi_{str}^{(l)}]
\longmapsto
(\alpha_{s,state}^{(l)},\,\alpha_{s,upd}^{(l)},\,\alpha_{v,upd}^{(l)})
}
\end{equation}
where $\Psi_{str}^{(l)}$ is the invariant structural statistic defined above.
SCDR is a bounded, permutation-invariant, geometry-preserving routing module for
semantic modulation of GVP dynamics.

\subsection{Equivariance and Stability Proofs}
\label{appendix:equivariance_stability}

We verify that the SCDR-modulated GVP block preserves the required geometric
symmetries and remains a bounded perturbation of the original backbone.

\paragraph{Proposition 1: equivariance preservation.}
Assume the base GVP block is permutation equivariant over nodes and
$SO(3)$-equivariant over vector channels under the reported reflection-sensitive
configuration. For any node permutation $\Pi$ and any proper rotation
$R\in SO(3)$, the base block satisfies
\begin{equation}
\begin{aligned}
F_s(\Pi\mathbf{s},\Pi\mathbf{V}R)&=\Pi F_s(\mathbf{s},\mathbf{V}),\\
F_V(\Pi\mathbf{s},\Pi\mathbf{V}R)&=\Pi F_V(\mathbf{s},\mathbf{V})R.
\end{aligned}
\end{equation}
Then the SCDR-modulated block preserves the same transformation laws.

\textit{Proof.}
SCDR reads vector channels only through norms:
\begin{equation}
\|\mathbf{V}_iR\|_2=\|\mathbf{V}_i\|_2.
\end{equation}
Each node descriptor $\mathbf{u}_i$ is therefore $SO(3)$-invariant. Since
$\Psi_{str}$ is built from permutation-invariant pooling operations
(mean, max, std, and count),
\begin{equation}
\Psi_{str}(\Pi\mathbf{s},\Pi\mathbf{V}R)=\Psi_{str}(\mathbf{s},\mathbf{V}).
\end{equation}
Consequently, the router latent and all routing gates are invariant:
\begin{equation}
\alpha(\Pi\mathbf{s},\Pi\mathbf{V}R)=\alpha(\mathbf{s},\mathbf{V}),
\end{equation}
for
\begin{equation}
\alpha\in\{\alpha_{s,state},\,\alpha_{s,upd},\,\alpha_{v,upd}\}.
\end{equation}

The AdaLN coefficients $\gamma(\mathbf{z}_{sem})$ and
$\beta(\mathbf{z}_{sem})$ depend only on the molecule-level semantic condition
and are broadcast to ligand nodes by the batch index. They remain invariant under
node permutation and under $SO(3)$ transformations, so scalar modulation preserves
permutation equivariance.
For the vector update, since $\alpha_{v,upd}$ is an invariant channel-wise
scalar broadcast over the Cartesian dimension,
\begin{equation}
\begin{aligned}
\Delta\hat{\mathbf{V}}(\Pi\mathbf{s},\Pi\mathbf{V}R)
&=\alpha_{v,upd}\odot\left(\Pi\Delta\mathbf{V}R\right)
\\
&=\Pi\left(\alpha_{v,upd}\odot\Delta\mathbf{V}\right)R
\\
&=\Pi\Delta\hat{\mathbf{V}}(\mathbf{s},\mathbf{V})R.
\end{aligned}
\end{equation}
Hence the SCDR-modulated block remains permutation equivariant and
$SO(3)$-equivariant under the reported configuration.

\paragraph{Proposition 2: bounded routing.}
SCDR yields a bounded local amplification of the original GVP update.

\textit{Proof.}
The dispatching nonlinearity is
\begin{equation}
b_a(x)=a\frac{x}{1+|x|}.
\end{equation}
This implies
\begin{equation}
|b_a(x)|<a.
\end{equation}
Since
\begin{equation}
\alpha=1+w_{\mathrm{prog}}(p)b_a(W\mathbf{q}),\,0\le w_{\mathrm{prog}}(p)\le1,
\end{equation}
we have
\begin{equation}
1-a<\alpha<1+a.
\end{equation}
The vector update is consequently bounded:
\begin{equation}
\|\Delta\hat{\mathbf{V}}\|
=\|\alpha_{v,upd}\odot\Delta\mathbf{V}\|\le(1+a_v)\|\Delta\mathbf{V}\|.
\end{equation}
The same argument applies to scalar updates, so each routed block remains a
finite multiplicative perturbation of the corresponding backbone update.

\paragraph{Proposition 3: initialization behavior.}
At initialization, the AdaLN state path recovers the scalar-normalized ligand
state of the replaced normalization path. The vector state is passed unchanged,
and either the warmup coefficient or the small-variance update initialization
keeps the residual gates near identity at startup.

\textit{Proof.}
The AdaLN projection is zero-initialized:
\begin{equation}
\gamma(\mathbf{z}_{sem})=0,\,\beta(\mathbf{z}_{sem})=0.
\end{equation}
At the start of an explicit training-progress warmup schedule,
\begin{equation}
w_{\mathrm{prog}}(p)=0.
\end{equation}
The routing gates then satisfy
\begin{equation}
\alpha_{s,state}=\alpha_{s,upd}=\alpha_{v,upd}=1.
\end{equation}
Substituting these values gives
\begin{equation}
\hat{\mathbf{s}}=\mathrm{LN}_s(\mathbf{s}),\,
\Delta\hat{\mathbf{s}}=\Delta\mathbf{s},\,
\Delta\hat{\mathbf{V}}=\Delta\mathbf{V}.
\end{equation}
The scalar AdaLN projection starts exactly from the scalar-normalized ligand
state, while the vector state path is unchanged. When the implementation runs
with active gates from the beginning, the gate heads are initialized with small
variance and zero bias; because $b_a(0)=0$ and $b_a$ is Lipschitz around the
origin, the multiplicative gates remain near one at startup. Proposition 2
controls their range throughout training.

Together, these propositions characterize SCDR as an invariant, bounded, and
stably initialized routing mechanism. It preserves the geometric structure of the
GVP backbone while enabling semantic modulation.

\subsection{Inference Consistency}
\label{appendix:inference_consistency}

Conditional sampling uses the transition families optimized during training and
keeps the same semantic-conditioning interface as the training dynamics. Before
the chain starts, the ligand size $N$, candidate edge set, categorical priors,
and semantic condition $\mathbf{z}_{sem}$ are fixed for the entire trajectory;
they are not changed by downstream docking, property scores, reference molecules,
or filtering. The generative chain starts from
\begin{equation}
\mathbf{X}_0\sim\mathcal{N}(\mathbf{c}_{\mathcal{P}}^{\,\mathrm{norm}},\mathbf{I}),\,
\mathbf{H}_0\sim\pi_0^H,\,
\mathbf{E}_0\sim\pi_0^E,
\end{equation}
in the same pocket-centered frame used during training. Let
$\delta_{\mathrm{sam}}=1/N_{\mathrm{sam}}$; the reported runs use
$N_{\mathrm{sam}}=500$. For
$k=0,\dots,N_{\mathrm{sam}}-1$ with $t_k=k/N_{\mathrm{sam}}$, the model forms
\begin{equation}
z_k=(\mathbf{X}_k,\mathbf{H}_k,\mathbf{E}_k),\,
\chi_k=(z_k,\mathcal{P},t_k,\mathbf{z}_{sem})
\end{equation}
and predicts
\begin{equation}
\widehat{\mathbf{v}}_\theta^X(\chi_k),
\widehat{\boldsymbol{\pi}}_k^H,
\widehat{\boldsymbol{\pi}}_k^E.
\end{equation}
At every solver step, $\mathbf{z}_{sem}$ is passed to the same SCDR-modulated GVP
parameterization used during training; coordinate and categorical components then
evolve according to their native transition rules.

For coordinates, the model is trained to predict the scaled velocity
\begin{equation}
\widehat{\mathbf{v}}_\theta^X\approx\frac{\mathbf{X}_1-\mathbf{X}_0}{c_x}.
\end{equation}
Euler sampling applies the inverse scaling:
\begin{equation}
\mathbf{X}_{k+1}=\mathbf{X}_k+\delta_{\mathrm{sam}}\,c_x\,\widehat{\mathbf{v}}_\theta^X(\chi_k).
\end{equation}
The coordinate sampler thereby integrates the velocity field learned by the
flow-matching objective.

For each discrete variable
\begin{equation}
\mathbf{Y}\in\{\mathbf{H},\mathbf{E}\},
\end{equation}
the terminal distribution for instance $i\in\mathcal{I}_Y$ has the softmax form
\begin{equation}
\widehat{\boldsymbol{\pi}}_{k,i}^Y=\mathrm{softmax}\left(f_{\theta,i}^Y(\chi_k)\right).
\end{equation}
The corresponding one-step bridge matrix is
\begin{equation}
\widehat{\mathbf{Q}}_{k+1\mid k,i}^{\theta}
=\beta_{k}\mathbf{I}+(1-\beta_k)\mathbf{1}(\widehat{\boldsymbol{\pi}}_{k,i}^Y)^\top,
\end{equation}
where
\begin{equation}
\beta_k=\frac{1-t_{k+1}}{1-t_k}.
\end{equation}
The next state is sampled as
\begin{equation}
\mathbf{Y}_{k+1,i}\sim
\mathrm{Cat}\left(\mathbf{Y}_{k,i}\widehat{\mathbf{Q}}_{k+1\mid k,i}^{\theta}\right).
\end{equation}

Inference therefore follows the same transition families used by the trainingobjectives. Coordinates use the learned flow-matching velocity; atom and bondupdates use learned terminal distributions inside the Markov bridge family. SCDRaffects inference only through the shared neural parameterization, leaving the continuous and categorical transition structures unchanged.

\subsection{Summary of the Verification}
\label{appendix:verification_summary}

The formulation positions LiFT as a structured heterogeneous
continuous--discrete generative model. Coordinates follow a flow-matching vector field, while atom and bond identities follow categorical Markov bridges. The semantic condition $\mathbf{z}_{sem}$ modulates the shared GVP parameterization while preserving these state spaces.

SCDR is formulated as an equivariant routing module with four preserved properties: permutation invariance, proper-rotation equivariance under the reported configuration, bounded local amplification, and stable initialization. Its DeepSets-motivated structural sensing provides an invariant summary of the current 3D molecular state, and its decoupled scalar gates inject semantic modulation through equivariant vector-channel scaling.

Overall, LiFT adds SCDR routing to the heterogeneous flow/bridge process while preserving boundedness, near-identity initialization, and equivariance. The appendix therefore establishes structural compatibility between semantic routing and the heterogeneous generative dynamics. Chemical validity, affinity gains, and semantic controllability are established empirically by the experiments and intervention ablations rather than by the symmetry proof alone.

\section{Extended Comparison with Related Work}
\label{appendix:related_comparison}

This appendix provides a structured comparison between LiFT and representative approaches spanning 1D--3D molecular generation, protein-conditioned sequence generation, LLM-assisted molecular design, and native pocket-conditioned 3D SBDD. The comparison is organized around neutral methodological dimensions---language input, target representation, the role of the 1D representation, native output, trainable components, interaction stage, and trajectory-time control---rather than a LiFT-specific ranking criterion. The goal is to clarify which interface each method addresses and where language- or sequence-derived information enters the molecular design pipeline.

\subsection{Systematic Methodological Comparison}
\label{appendix:systematic_comparison}

Table~\ref{tab:systematic_related_comparison} summarizes the main methodological distinctions. Existing approaches cover several complementary interfaces. NExT-Mol \citep{liu2025nextmol3ddiffusionmeets} and MolSculpt \citep{chen2025molsculptsculpting3dmolecular} connect molecular-language representations with target-free 3D molecular generation. MolChord \citep{zhang2025molchord} and TamGen \citep{Wu2024TamGenDD} use protein information to condition molecular-sequence generation, with the generated molecular string serving as the primary molecular proposal. ELILLM \citep{hu2026empoweringllmsstructurebaseddrug} searches a pretrained LLM latent space using target-dependent optimization feedback, while CIDD \citep{gao2025cidd} applies LLM-based refinement after an initial SBDD generation stage. Native 3D SBDD methods such as AR \citep{luo20223dgenerativemodelstructurebased}, Pocket2Mol \citep{pmlr-v162-peng22b}, TargetDiff \citep{guan20233dequivariantdiffusiontargetaware}, DecompDiff \citep{guan2024decompdiffdiffusionmodelsdecomposed}, PAFlow \citep{zhou2025priorguided}, and DrugFlow \citep{schneuing2025multidomain} directly model pocket-conditioned ligand geometry without an open-ended language-derived semantic condition.

LiFT occupies a different interface within this landscape: open-ended design preferences and pocket-derived specifications are converted into a SMILES-derived soft semantic prior rather than a final ligand, while the final molecular topology and pose are generated in the native pocket-conditioned 3D space. The semantic condition acts inside the flow trajectory, and SCDR adjusts its influence according to flow time and the evolving ligand state. This comparison concerns the location and role of cross-modal conditioning; it does not imply that the individual components used by LiFT are themselves new molecular or geometric primitives.

\begin{table*}[t]
\centering
\caption{Systematic methodological comparison of LiFT with representative related approaches. The dimensions describe where language, molecular strings, protein information, and 3D generation interact. ``Native output'' denotes the direct output of the generative model before optional post-hoc conformer construction, docking, or refinement.}
\label{tab:systematic_related_comparison}
\setlength{\tabcolsep}{3.0pt}
\renewcommand{\arraystretch}{1.10}
\resizebox{\textwidth}{!}{
\begin{tabular}{p{3.30cm} p{2.51cm} p{2.06cm} p{2.79cm} p{2.33cm} p{2.88cm} p{2.42cm} p{2.41cm}}

\toprule
\textbf{Method} &
\textbf{Language Input} &
\textbf{Target Representation} &
\textbf{Role of 1D Representation} &
\textbf{Native Output} &
\textbf{Trainable Components / Objective} &
\textbf{Interaction Stage} &
\textbf{Trajectory-Time Control} \\
\midrule

\textbf{LiFT} &
Open-ended task preferences, pocket descriptions, property trade-offs, and optional references &
Native 3D protein pocket &
Intermediate SMILES-derived \textbf{soft semantic prior}, not the final ligand &
Pocket-conditioned 3D ligand and aligned pose &
Lightweight semantic projector and SCDR are optimized under the existing flow-matching objective while retaining the standard geometric backbone &
Semantic information acts inside the 3D flow trajectory &
Dynamic routing according to flow time and the evolving 3D ligand state \\

NExT-Mol \citep{liu2025nextmol3ddiffusionmeets} &
No open-ended human design instruction; molecular string is provided &
No protein target &
SELFIES specifies the molecular identity to be reconstructed in 3D &
A conformer of the specified molecule &
1D--3D representation alignment and diffusion-based conformer generation &
Molecular-string information is aligned with the corresponding 3D molecule during conformer generation &
Fixed molecule-level alignment; no pocket-conditioned semantic routing \\

MolSculpt \citep{chen2025molsculptsculpting3dmolecular} &
No open-ended task-language interface &
No protein target &
SELFIES-derived molecular-language features provide generic chemical knowledge &
Target-free 3D molecule &
Learnable queries/projector and a 3D diffusion objective &
Molecular-language features condition 3D diffusion &
No target-pocket state or language-guided state-dependent routing \\

MolChord \citep{zhang2025molchord} &
Text-aligned molecular representations, but not open-ended LiFT-style task control &
Protein structure representation &
Autoregressively decoded molecular string is the final molecular proposal &
SMILES; 3D conformer and docking are obtained afterward &
Structure adapter and molecular sequence decoder trained with a sequence-generation objective &
Protein representation conditions 1D decoding &
No native 3D generation trajectory \\

TamGen \citep{Wu2024TamGenDD} &
No open-ended human design-language interface &
Protein or pocket representation &
Generated SMILES is the final molecular proposal &
SMILES; conformer construction and docking are post hoc &
Target-conditioned chemical language model with an autoregressive objective &
Target information conditions sequence decoding &
No native 3D generation trajectory \\

ELILLM \citep{hu2026empoweringllmsstructurebaseddrug} &
Optimization objectives rather than an open-ended 3D design interface &
Target enters primarily through a docking oracle &
Optimized latent is decoded into the final SMILES candidate &
SMILES; conformer generation and docking follow &
Bayesian optimization in a pretrained LLM latent space &
Control occurs during latent-space exploration before 3D evaluation &
No interaction with an evolving 3D ligand state \\

CIDD \citep{gao2025cidd} &
LLM refinement instructions informed by interaction analysis &
Protein pocket and docking-derived feedback &
The molecular string represents an already generated candidate to be edited &
Refined molecular proposal after an initial SBDD stage &
Prompted or iterative LLM refinement using docking/interaction feedback &
Language intervenes after initial molecule generation &
Post-generation refinement rather than trajectory-internal control \\

AR / Pocket2Mol / TargetDiff / DecompDiff / PAFlow / DrugFlow
\citep{luo20223dgenerativemodelstructurebased,pmlr-v162-peng22b,guan20233dequivariantdiffusiontargetaware,guan2024decompdiffdiffusionmodelsdecomposed,zhou2025priorguided,schneuing2025multidomain} &
No open-ended human-language instruction &
Native 3D protein pocket &
Usually no language-derived 1D semantic condition &
Pocket-conditioned 3D ligand and pose &
Autoregressive-, diffusion-, or flow-based 3D generation objectives &
Pocket geometry participates throughout native 3D generation &
State-dependent geometric generation, but no language-derived semantic routing \\

\bottomrule
\end{tabular}}
\end{table*}

\subsection{Focused Comparison with ELILLM}
\label{appendix:ELILLM_comparison}

ELILLM \citep{hu2026empoweringllmsstructurebaseddrug} is particularly relevant because both methods connect language-model representations with structure-based molecular design, but the representations play different roles and act at different stages. Table~\ref{tab:ELILLM_lift_comparison} therefore separates the two pipelines along the dimensions most directly related to their problem formulation.

\begin{table*}[t]
\centering
\caption{Focused methodological comparison between ELILLM and LiFT. The distinction is primarily the role of the language-derived representation and the space in which target-aware control occurs.}
\label{tab:ELILLM_lift_comparison}
\setlength{\tabcolsep}{5pt}
\renewcommand{\arraystretch}{1.10}
\resizebox{\textwidth}{!}{
\begin{tabular}{p{4.0cm} p{5.7cm} p{5.7cm}}
\toprule
\textbf{Comparison Dimension} &
\textbf{ELILLM \citep{hu2026empoweringllmsstructurebaseddrug}} &
\textbf{LiFT} \\
\midrule

Core problem &
Searches for molecular candidates with improved docking objectives in a pretrained LLM latent space &
Enables language design intent to participate in native pocket-conditioned 3D generation \\

Role of language/latent representation &
Optimized and decoded into the final SMILES candidate &
Forms a chemical-semantic prior that is not the final output \\

How target information enters &
Primarily through a docking oracle and Bayesian-optimization feedback &
The 3D pocket directly conditions the flow backbone and is also interpreted by the agent to form a pocket-derived specification \\

Native generation space &
LLM latent / 1D sequence space &
3D ligand state space in the target-pocket coordinate frame \\

Native output &
SMILES &
Atom types, 3D coordinates, and a pocket-aligned pose \\

How the 3D structure is obtained &
Conformer embedding, optimization, and docking after sequence generation &
Native generation within a pocket-conditioned flow trajectory \\

Where control occurs &
Before generation, through search in the LLM latent space &
During generation, by dynamically modulating the velocity field according to the intermediate 3D state \\

State-aware 3D routing &
No &
\textbf{Yes, through SCDR} \\

Adaptation to a new task &
Changes the optimization objective and reruns latent-space search &
Changes the inference-time language instruction without retraining the generator \\

\bottomrule
\end{tabular}}
\end{table*}

In short, ELILLM searches and optimizes a final 1D candidate in an LLM latent space and subsequently obtains its 3D evaluation through conventional conformer construction and docking. LiFT instead uses language-derived chemical semantics as a soft condition that directly operates inside a native pocket-conditioned 3D generation trajectory. Accordingly, the two methods address related SBDD goals through different interfaces: ELILLM focuses on molecular-candidate optimization in language-model latent space, whereas LiFT focuses on state-aware cross-modal conditioning of an evolving 3D generator.

\end{document}